\documentclass{article}

\usepackage{array}
\usepackage{subcaption}
\usepackage{amsmath} 
\usepackage{stmaryrd}

\usepackage[nonatbib,preprint]{neurips_2024}

\usepackage[utf8]{inputenc} 
\usepackage[T1]{fontenc}    
\usepackage{hyperref}       
\usepackage{url}            
\usepackage{booktabs}       
\usepackage{amsfonts}       
\usepackage{nicefrac}       
\usepackage{microtype}      
\usepackage{xcolor}         
\usepackage{times}
\usepackage{epsfig}
\usepackage{graphicx}
\usepackage{amsmath}
\usepackage{amssymb}
\usepackage{bbm}
\usepackage{multirow}
\usepackage{color}
\usepackage{colortbl}
\usepackage{xcolor}
\usepackage{multirow}
\usepackage{makecell}
\usepackage{tikz}
\usepackage{siunitx}
\usepackage{bbding}
\usepackage[capitalize]{cleveref}
\usepackage[normalem]{ulem}
\usepackage{expl3}
\ExplSyntaxOn
\newcommand\latinabbrev[1]{
	\peek_meaning:NTF . {
		#1\@}%
	{ \peek_catcode:NTF a {
			#1.\@ }%
		{#1.\@}}}
\ExplSyntaxOff

\definecolor{applegreen}{rgb}{0.0, 0.5, 0.0}

\definecolor{aliceblue}{rgb}{0.94, 0.97, 1.0}

\title{Deep-PE: A Learning-Based Pose Evaluator for \\
Point Cloud Registration}

\author{
    Junjie Gao \\
    Shandong University\\
    \texttt{Junjie.gao95.m@gmail.com}
  \And
    Chongjian Wang \\
    Shandong University of Science and Technology\\
    \texttt{202311080223@sdust.edu.cn}
  \And
    Zhongjun Ding \\
    National Deep Sea Center\\
    \texttt{dzj@ndsc.org.cn}
    \And
    Shuangmin Chen\thanks{Corresponding author.}\\
    Qingdao University of Science and Technology\\
    \texttt{csmqq@163.com}
    \AND
    Shiqing Xin \\
    Shandong University\\
    \texttt{xinshiqing@sdu.edu.cn}
    \And
    Changhe Tu \\
    Shandong University \\
    \texttt{chtu@sdu.edu.cn}
    \And
    Wenping Wang \\
    Texas A\&M University \\
    \texttt{wenping@tamu.edu}
}

\begin{document}

\maketitle

\begin{abstract}
In the realm of point cloud registration, the most prevalent pose evaluation approaches are statistics-based, identifying the optimal transformation by maximizing the number of consistent correspondences. However, registration recall decreases significantly when point clouds exhibit a low overlap rate, despite efforts in designing feature descriptors and establishing correspondences.
In this paper, we introduce Deep-PE, a lightweight, learning-based pose evaluator designed to enhance the accuracy of pose selection, especially in challenging point cloud scenarios with low overlap. Our network incorporates a Pose-Aware Attention (PAA) module to simulate and learn the alignment status of point clouds under various candidate poses, alongside a Pose Confidence Prediction (PCP) module that predicts the likelihood of successful registration. 
These two modules facilitate the learning of both local and global alignment priors.
Extensive tests across multiple benchmarks confirm the effectiveness of Deep-PE. Notably, on 3DLoMatch with a low overlap rate, Deep-PE significantly outperforms state-of-the-art methods by at least 8\% and 11\% in registration recall under handcrafted FPFH and learning-based FCGF descriptors, respectively. To the best of our knowledge, this is the first study to utilize deep learning to select the optimal pose without the explicit need for input correspondences.

\end{abstract}


\section{Introduction}


Finding an accurate rigid transformation between two unaligned partial point clouds—a process known as point cloud registration—is a fundamental task in the fields of graphics, vision, and robotics. This task becomes increasingly challenging with low-overlap point clouds. In response to this, a significant number of advanced feature descriptors \cite{huang2021predator, Yu2021CoFiNetRC, Qin2022GeometricTF} and robust pose estimators \cite{chen2022sc2, jiang2023robust, zhang2023MAC} have been developed. However, recent progress in registration seems to have encountered a bottleneck, especially in improving the registration recall in low-overlap scenes.

\begin{figure}[h]
    \centering
    \begin{subfigure}{0.49\linewidth}
        \centering
        \includegraphics[width=\linewidth]{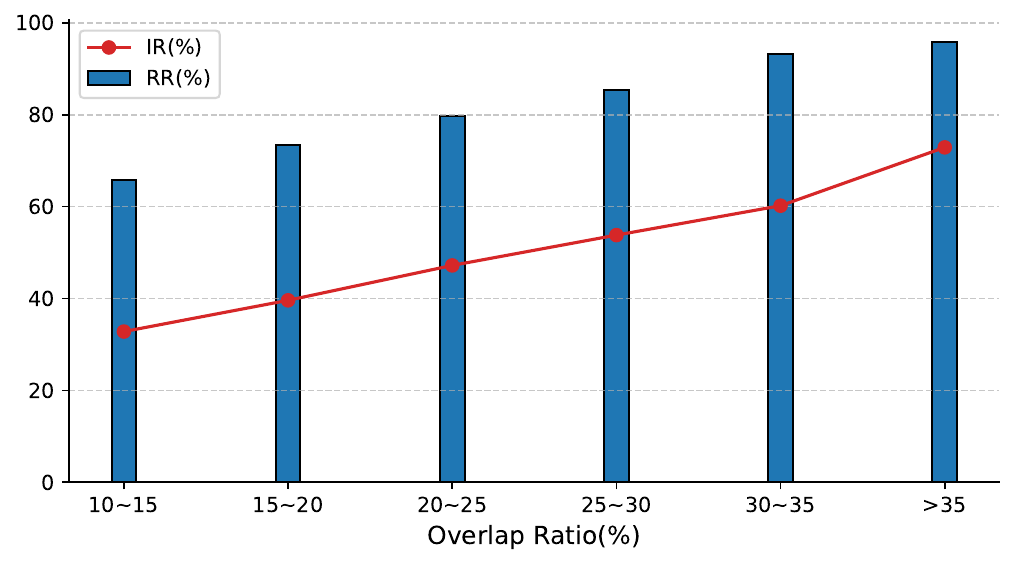}
        \vspace{-18pt}
        \caption{}
        \label{fig1a}
    \end{subfigure}
    \hspace{0.001\linewidth}
    \begin{subfigure}{0.49\linewidth}
        \centering
        \includegraphics[width=\linewidth]{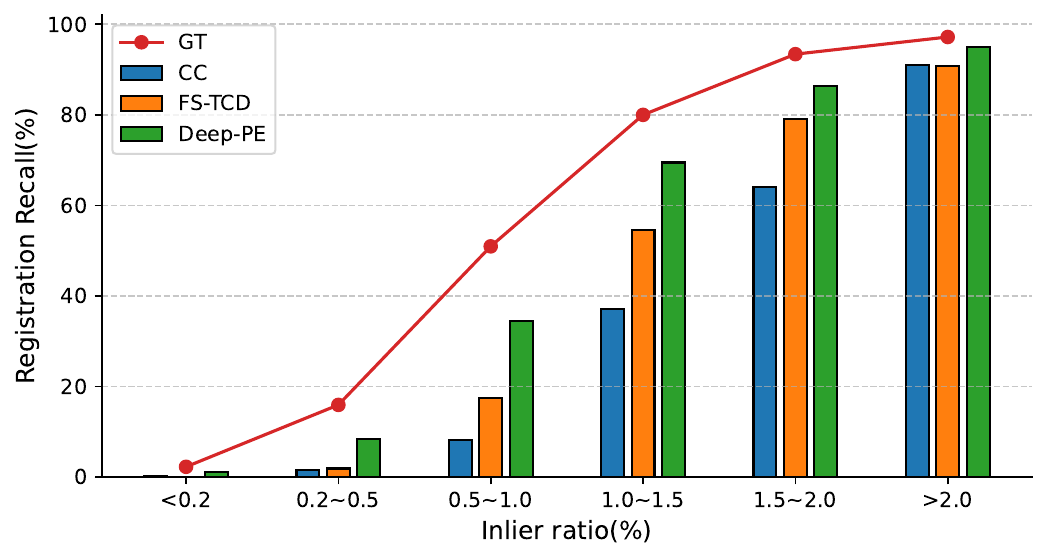}
        \vspace{-18pt}
        \caption{}
        \label{fig1b}
    \end{subfigure}
    \vspace{-5pt}
    \caption{
    (a) Even in the well-known registration pipeline \cite{Qin2022GeometricTF}, the Inlier Ratio (IR) and Registration Recall (RR) deteriorates rapidly for point cloud pairs with an overlap of <30\%.
    (b) In low-overlap 3DLoMatch benchmark, compared to the statistics-based pose evaluators, CC \cite{Fischler1981RandomSC} and FS-TCD \cite{chen2023sc}, Deep-PE demonstrates a significant advantage in cases of low inlier ratios and also approaches the ground-truth value as the inlier ratio increases.
    }
    \label{fig1}
\end{figure}

As shown in \cref{fig1a}, we found that even in the well-known registration pipeline \cite{Qin2022GeometricTF}, when the overlap ratio drops below 30\%, the inlier ratio (IR) of extracted correspondences rapidly decreases, leading to a simultaneous decrease in registration recall (RR). To explore this phenomenon further, we extracted correspondences using the FPFH descriptor \cite{Rusu2009FastPF} and generated candidate poses using SC2-PCR \cite{chen2022sc2} on the low-overlap 3DLoMatch dataset. Subsequently, based on the inlier ratio, we divided all point cloud pairs into six different groups. As illustrated in \cref{fig1b}, it can be seen that although the pose estimator can generate a set of candidate poses containing the correct transformation (Ground Truth, GT). However, during the process of selecting the optimal pose, the currently most commonly used and advanced statistics-based pose evaluators, CC \cite{Fischler1981RandomSC} and FS-TCD \cite{chen2023sc}, exhibit varying degrees of performance degradation when the inlier ratio falls below 1\%, largely due to their excessive reliance on the quality of input correspondences.


Motivated by this observation, we propose a lightweight deep learning-based pose evaluator named Deep-PE to enhance the accuracy of pose selection. Unlike traditional pose evaluators, Deep-PE fully integrates alignment priors, endowing it with global pose perception capabilities and making it sensitive solely to pose without dependence on the quality of input correspondences. Consequently, as demonstrated in \cref{fig1b}, our method shows stronger robustness against low inlier ratios. Additionally, unlike existing pose evaluators \cite{Fischler1981RandomSC, Yang2021TowardEA, chen2023sc} that must select one of the candidate poses as the final transformation, Deep-PE predicts with a confidence score, enabling it to identify registration failures effectively.





To summarize, the main contributions are as follows:
\begin{itemize}
\item A pioneering learning-based pose evaluator is proposed, which does not rely on the quality of input correspondences and has the capability to rectify registration failures.

\item A pose-aware attention module is proposed to simulate and learn the alignment status of point clouds under various candidate poses, and a pose confidence prediction module is proposed to predict the likelihood of successful registration.

\item  A weighted cross-entropy loss is proposed, which reweights the loss of each candidate pose according to its similarity with the ground truth  for better identification of the optimal transformation.

\item  A series of experimental results demonstrate that  our proposed method performs favorably against state-of-the-art pose evaluators.

\end{itemize}

\section{Related Work}
\subsection{Estimator-based Registration}
The registration pipeline of such methods typically includes three steps: firstly, feature descriptors \cite{Tombari2010UniqueSO,Rusu2009FastPF,Gojcic2018ThePM,Choy2019FullyCG,Ao2020SpinNetLA,Wang2023RoRegPP,Yu2021CoFiNetRC,Qin2022GeometricTF,Bai2020D3FeatJL,huang2021predator,lu2019deepvcp} are meticulously designed to describe the input point clouds for extracting correspondences; secondly, robust pose estimators \cite{Fischler1981RandomSC, Barth2017GraphCutR,Leordeanu2005AST, Choy2020DeepGR, Bai2021PointDSCRP, Pais20193DRegNetAD, chen2023sc, zhang2023MAC,jiang2023robust} generate a series of candidate poses; finally, an effective pose evaluator \cite{Fischler1981RandomSC,Yang2021TowardEA,chen2023sc} is used to select the optimal pose from them as the final transformation.

For the methods involved in the first step, they can be divided into two categories based on how they extract correspondences. The first category aims to detect more repeatable keypoints \cite{Bai2020D3FeatJL,huang2021predator} and learn more powerful keypoint descriptors \cite{lu2019deepvcp}. The second category considers all possible correspondences without keypoint detection \cite{Tombari2010UniqueSO,Rusu2009FastPF,Gojcic2018ThePM,Choy2019FullyCG,Ao2020SpinNetLA,Wang2023RoRegPP,Yu2021CoFiNetRC,Qin2022GeometricTF}. When extracting correspondences between two point clouds, non-overlapping points can be entirely regarded as interference factors. This leads to the inability of these two categories of methods to achieve satisfactory performance on low-overlap point clouds, despite their excellent performance in high-overlap scenarios.

For the pose estimator mentioned in the second step, significant progress has been made in recent years, and it can be broadly categorized into two categories: traditional \cite{Fischler1981RandomSC, Barth2017GraphCutR,Leordeanu2005AST, chen2023sc, zhang2023MAC} and learning-based \cite{Choy2020DeepGR,Bai2021PointDSCRP, Pais20193DRegNetAD}. Traditional pose estimators typically obtain outlier-free subsets for pose estimation through extensive iterations \cite{Fischler1981RandomSC, Barth2017GraphCutR,Myatt2002NAPSACHN} or checking compatibility \cite{Leordeanu2005AST, chen2022sc2} between correspondences. Learning-based pose estimators \cite{Choy2020DeepGR,Pais20193DRegNetAD} typically consist of a classification network to reject outliers and a regression network to compute transformations.

As the final step in the registration pipeline,  the pose evaluation has been greatly neglected, to the best of our knowledge, all current pose evaluators \cite{Fischler1981RandomSC, Yang2021TowardEA, chen2023sc} are statistics-based, assessing the alignment quality of candidate poses by  maximizing
the number of consistent correspondence. In high-overlap scenarios, a high inlier ratio typically supports this evaluation mechanism. However, low-overlap point clouds often come with a much lower inlier ratio, rendering this reliance on the quality of input correspondences ineffective. In contrast, our proposed method fully learns the alignment status of point clouds under various candidate
poses and predicts the likelihood of successful registration in terms of confidence. Compared to traditional statistics-based pose evaluators, our approach offers higher selecting precision and the ability to identify registration failures.

\subsection{Estimator-free Registration}
Some efforts have diverged from the traditional approach of combining learned descriptors with robust pose estimators, choosing instead to fully integrate the pose estimation process within their training pipeline \cite{Wang2019DeepCP,Yuan2020DeepGMRLL,Li2021PointNetLKR,Huang2020FeatureMetricRA, Yew2020RPMNetRP, mei2023unsupervised}. These methods can be further classified into three categories. The first category \cite{Wang2019DeepCP, Yew2020RPMNetRP} follows the ICP approach \cite{Besl1992AMF}, iteratively establishing soft correspondences and computing transformations using a differentiable weighted SVD \cite{Papadopoulo2000EstimatingTJ}. The second category \cite{Li2021PointNetLKR,Huang2020FeatureMetricRA} first extracts a global feature vector \cite{Qi2016PointNetDL}  for each point cloud and then regresses transformations using these feature vectors. The final category is distribution-level methods \cite{Yuan2020DeepGMRLL, mei2023unsupervised}, which model point clouds as probability distributions (e.g., Gaussian mixture models \cite{jian2010robust}) and perform alignment using correlation-based optimization frameworks. Although estimator-free registration pipelines avoid some issues introduced by pose estimators and achieved promising results on single synthetic shapes, they could fail in large-scale scenes and are unable to handle more challenging  scenarios such as low-overlap point clouds. We find that combining advanced feature descriptors with a robust pose estimator is sufficient to generate a set of candidate poses containing the correct transformation. Therefore, in the network design, we fully leverage the advantages of an estimator-free registration pipeline, opting for the network to select the optimal ones from candidate poses instead of directly regressing pose parameters. This strategy significantly reduces the training difficulty of the network.

\subsection{Transformer in Vision}
Convolutional Neural Networks (CNNs) excel in local feature description, but are limited by their receptive field size, whereas Transformer networks \cite{Vaswani2017AttentionIA} are better at capturing long-range contextual information and facilitating effective information exchange between inputs. Therefore, these two methods complement each other and are widely applied in 2D and 3D feature matching tasks \cite{Wang2019SuperGLUEAS, Sun2021LoFTRDL, huang2021predator, Yu2021CoFiNetRC, Qin2022GeometricTF}. However, Transformers are computationally expensive and typically can only be used at low-resolution levels.  With the development of deep learning, numerous neural network frameworks \cite{Qi2016PointNetDL, Thomas2019KPConvFA, Choy2019FullyCG, wang2019dynamic} have emerged, but enabling the network to learn the alignment status under different poses is crucial for our task. Recently, we found that the Epipolar Transformer \cite{he2020epipolar} with constrained attention regions has achieved remarkable success in 2D stereo matching tasks. Inspired by this, we propose a lightweight pose-aware attention module. The module can flexibly adjust attention regions based on different candidate poses, and it is local and limited. Therefore, on the one hand, it can simulate and learn the alignment status of point clouds under various candidate poses, while on the other hand, it significantly reduces computational costs. Then, by combining it with the pose confidence prediction module, we can predicts the likelihood of successful registration for each candidate pose.

\section{Method}
\subsection{Insight}

\begin{figure}[h]
  \centering
  \includegraphics[width=.9\linewidth]{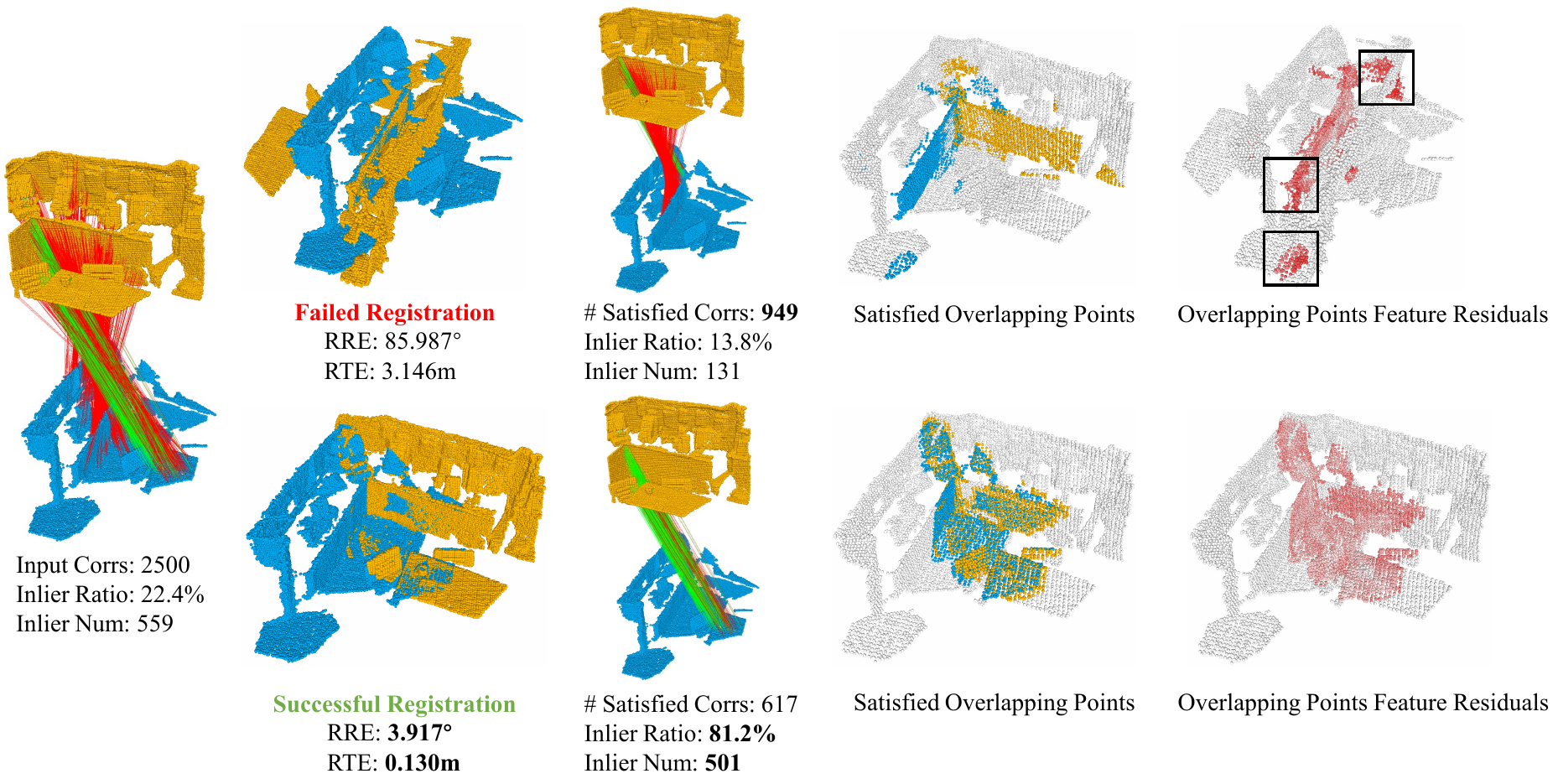}
  \vspace{-5pt}
  \caption{
    Illustration of our insight: Incorrect poses frequently align areas in the point clouds that should not be aligned; points within these regions typically display larger feature residuals. The fourth column depicts the overlapping regions predicted under the ground truth pose, while the fifth column illustrates the severity of feature residuals, with red colors denoting larger residuals. RRE: Relative Rotation Error. RTE: Relative Translation Error.
    }
  \label{fig.2}
\end{figure}

Our research builds on the following observation: As depicted in \cref{fig.2}, when the inlier ratio of input correspondences is low, overly relying on maximizing consistent correspondences for pose estimation is often ineffective. Incorrect poses result in large feature residuals within overlapping regions (see highlighted windows).

\subsection{Overview}
\begin{figure}[h]
  \centering
  \vspace{-12pt}
  \includegraphics[width=\linewidth]{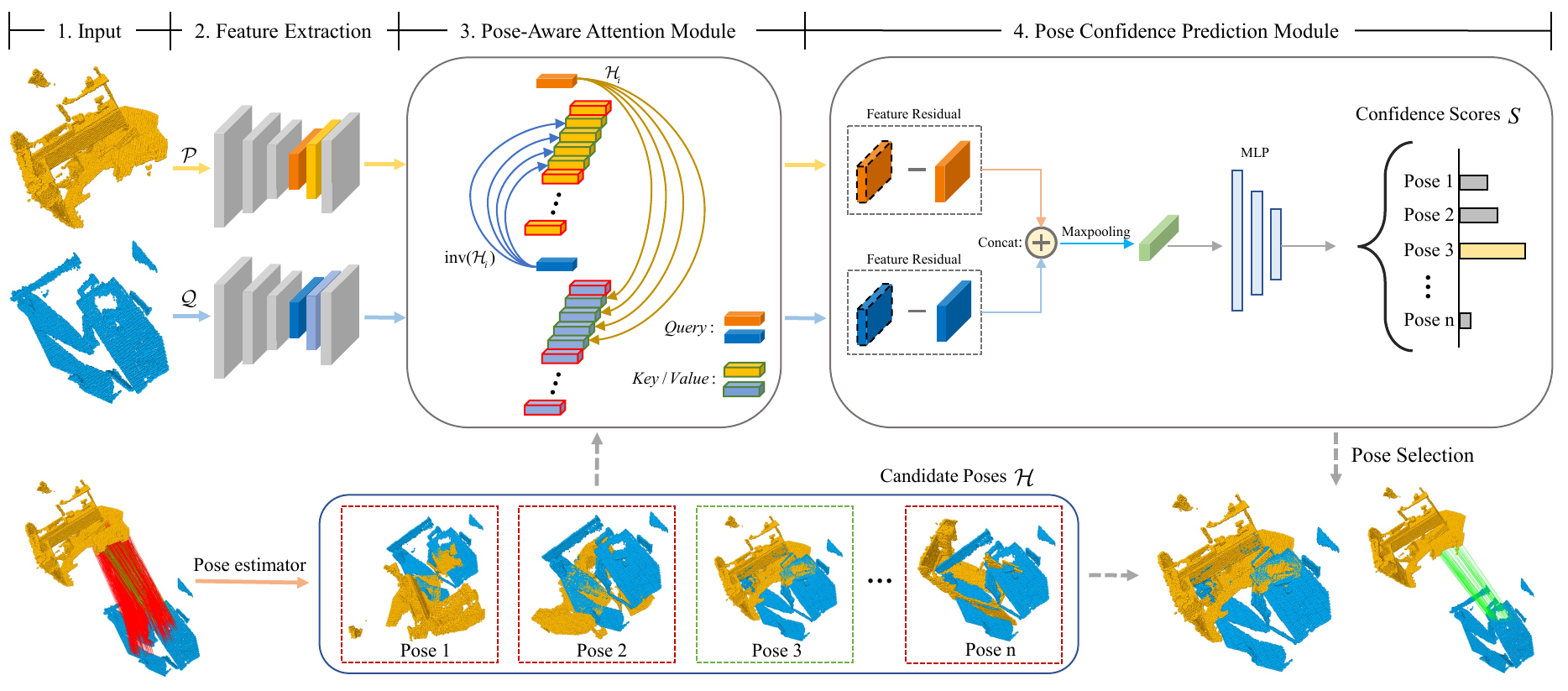}
  \vspace{-12pt}
  \caption{
  Deep-PE mainly consists of three components:
  1. The pre-trained feature extractor downsamples the input point clouds and learns features in multiple resolution levels. The points and features of the coarsest and penultimate layers are cached and reused for each candidate pose $\mathcal{H}_i$.
  2. The pose-aware attention module adjusts attention regions based on different candidate poses and embeds relevant features into coarse-level features.
  3. The pose confidence prediction module first calculates the feature residuals before and after coarse-level feature updates for each point cloud, then combines them through concatenation and max-pooling operations. Finally, a Multilayer Perceptron (MLP) layer is employed to predict the confidence score  $\mathcal{S}_i$ associated with each candidate pose $\mathcal{H}_i$. Then, the pose with the maximum confidence score is selected as the final transformation.
  }
  \label{fig.3}
\end{figure}

Given two point clouds: $\mathcal{P}=\left\{\mathbf{p}_i\in \mathbb{R}^{3} \mid i=1, \ldots, N\right\}$ and $\mathcal{Q}=\left\{\mathbf{q}_j\in \mathbb{R}^{3} \mid j=1, \ldots, M\right\}$, and a set of candidate poses $\mathcal{H} \in \{\mathbf{R}, \mathbf{t}\}$ generated by the pose estimator, Deep-PE processes each candidate pose $\mathcal{H}_i$ and  predicts the relevant confidence score $\mathcal{S}_i$, which describes the alignment quality of the input point clouds. As shown in Fig.~\ref{fig.3}, Deep-PE mainly consists of three components: a Feature Extraction (FE) module, a Pose-Aware Attention (PAA) module, and a Pose Confidence Prediction (PCP) module. The last two modules are our main design.


\subsection{Feature Extractor}
Due to the typically large number of candidate poses generated by pose estimators, a lightweight network design is essential to ensure efficient batch processing. Additionally, our primary focus is on the training aspects of pose evaluation rather than feature extraction. To reduce training complexity and improve model convergence speed, we choose to utilize the pre-trained Geotransformer \cite{Qin2022GeometricTF} as the feature extractor, which employs KPConv \cite{Thomas2019KPConvFA} to downsample the input point cloud and simultaneously extract point-wise features. As depicted in \cref{fig.3}, we only utilize two layers of points and their corresponding features. To further enhance feature learning and facilitate subsequent computations, we employ two linear layers to map these features to the same dimension. Specifically, the points and their updated features at the coarsest layers are represented as $\hat{\mathcal{P}}$/$\hat{\mathcal{Q}}$ and $\hat{\mathbf{F}}^{\mathcal{P}}$/$\hat{\mathbf{F}}^{\mathcal{Q}} \in \mathbb{R}^{|\hat{\mathcal{P}}/\hat{\mathcal{Q}}| \times d}$, while those at the penultimate layers are represented as $\Tilde{\mathcal{P}}$/$\Tilde{\mathcal{Q}}$ and $\Tilde{\mathbf{F}}^{\mathcal{P}}$/$\Tilde{\mathbf{F}}^{\mathcal{Q}} \in \mathbb{R}^{|\Tilde{\mathcal{P}}/\Tilde{\mathcal{Q}}| \times d}$.

\subsection{Pose-Aware Attention Module}\label{sec3.4}
Given that Deep-PE needs to predict the confidence of each candidate pose, we choose to cache and reuse the obtained features. This design not only reduces computational costs but also avoids potential adverse effects on training caused by feature divergence resulting from each prediction.

To enable the network to learn and evaluate the alignment quality of different candidate poses, we simulate the alignment status of point clouds by adjusting the attention region. Specifically, under the current candidate pose $\mathcal{H}_i$, each point $\hat{\mathbf{p}}_i$ in $\hat{\mathcal{P}}$ with associated feature $\hat{\mathbf{F}}^{\mathcal{P}}_i$ has a corresponding point defined as $\hat{\mathbf{p}}_i'=\mathcal{H}_i \hat{\mathbf{p}}_i$, and then we adopt a $k$-nearest neighbor search within $\Tilde{\mathcal{Q}}$ to produce $k$ points $\mathcal{K}^{\Tilde{\mathcal{Q}}}_{\mathbf{p}_i}$. The associated learned features are denoted as $\mathcal{F}^{\Tilde{\mathcal{Q}}}_{\mathbf{p}_i} \in \mathbb{R}^{k \times d}$. If the distance from the sampling points to $\hat{\mathbf{p}}_i'$ exceeds the threshold $t$, we zero-pad the features. Thus, we build a feature volume ${\mathcal{F}}^{\hat{\mathcal{P}} \rightarrow  \Tilde{\mathcal{Q}}} \in \mathbb{R}^{|\hat{\mathcal{P}}| \times k \times d}$. The feature volume ${\mathcal{F}}^{\hat{\mathcal{Q}} \rightarrow  \Tilde{\mathcal{P}}}$ for $\hat{\mathcal{Q}}$ is computed and denoted similarly.

Accordingly, given the coarse point features $\hat{\mathbf{F}}^{\mathcal{P}}$ and the relevant feature volume ${\mathcal{F}}^{\Tilde{\mathcal{P}} \rightarrow  \Tilde{\mathcal{Q}}}$, the pose-aware attention feature $\hat{\mathbf{H}}^{{\mathcal{P}}} \in \mathbb{R}^{|\hat{\mathcal{P}}| \times d}$ is the weighted sum of all projected feature volumes:
\begin{equation}
    \hat{\mathbf{h}}_i^{\mathcal{P}}=\sum_{j=1}^{k} a_{i, j}\left(\mathbf{f}_j \mathbf{W}^V\right)
\end{equation}
where $\mathbf{f}_j \in \bar{\mathcal{F}}^{\Tilde{\mathcal{P}} \rightarrow  \Tilde{\mathcal{Q}}}$, and the weight coefficient $a_{i, j}$ is computed by a row-wise softmax on the attention score $e_{i, j}$, with $e_{i, j}$ computed as:
\begin{equation}
e_{i, j}=\frac{\left(\mathbf{f}_i 
 \mathbf{W}^Q\right)\left(\mathbf{f}_j \mathbf{W}^K\right)^T}{\sqrt{d}}
\end{equation}
Here, $\mathbf{f}_i \in \hat{\mathbf{F}}^{\mathcal{P}}$, and $\mathbf{W}^Q, \mathbf{W}^K, \mathbf{W}^V \in \mathbb{R}^{d \times d}$ are the respective projection matrices for queries, keys, and values. The pose-aware attention features $\hat{\mathbf{H}}^{{\mathcal{Q}}}$ for $\hat{\mathbf{F}}^{\mathcal{Q}}$ are computed in the same manner.

\subsection{Pose Confidence Prediction Module}

\begin{figure}[t]
  \centering
  \includegraphics[width=.8\linewidth]{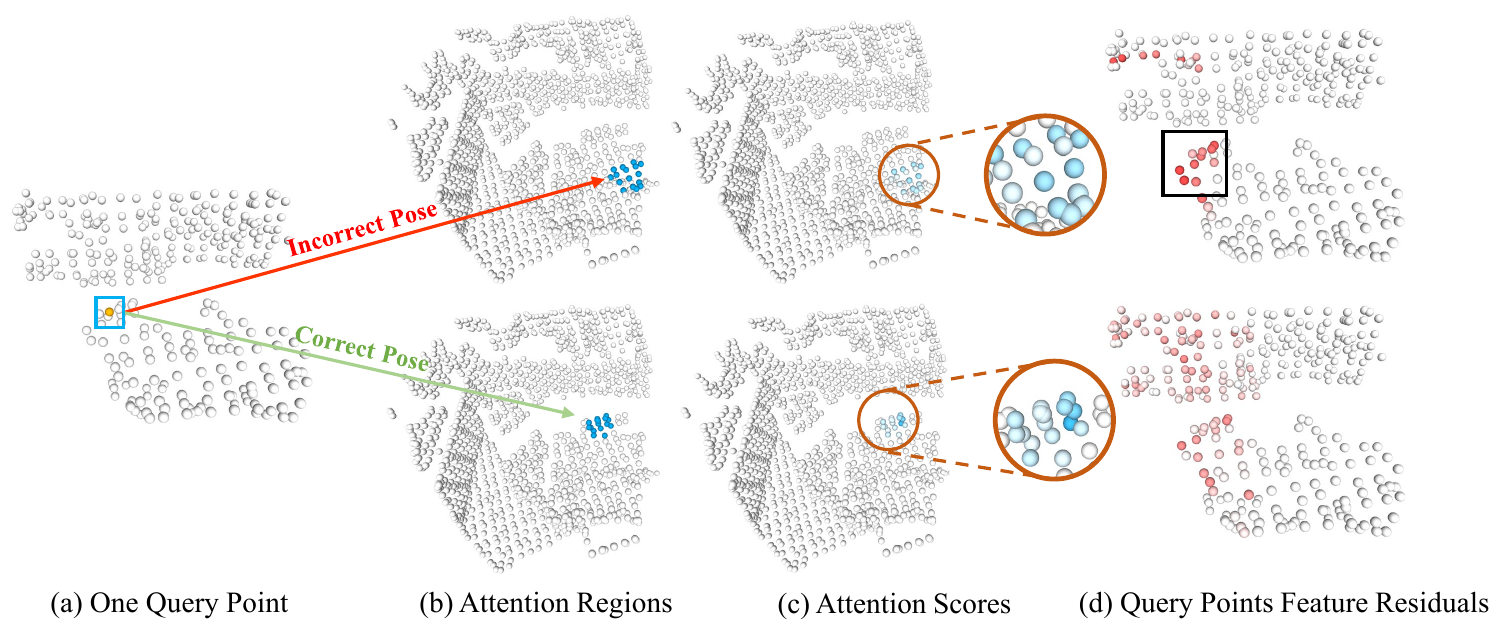}
  \vspace{-5pt}
  \caption{
  Illustration of attention regions and scores under correct and incorrect poses. It's evident that, with the correct pose, attention scores are higher in regions resembling the query point, leading to smaller feature residuals before and after updating. Conversely, the opposite holds true under an incorrect pose.
  }
  \label{fig.4}
\end{figure}

After passing through the pose-aware attention module, the coarse point features $\hat{\mathbf{F}}^{\mathcal{P}}$/$\hat{\mathbf{F}}^{\mathcal{Q}}$ have been updated to $\hat{\mathbf{H}}^{\mathcal{P}}$/$\hat{\mathbf{H}}^{\mathcal{Q}}$. As illustrated in \cref{fig.4}, notable disparities between the features updated with correct and incorrect poses are observed. The primary reason lies in the attention mechanism's ability to assign greater weight to similar features. In this scenario, when the pose is correct, the corresponding nearest neighbor features tend to exhibit some similarities, and vice versa.

As depicted in \cref{fig.3}, to discern the variance in feature updates across different poses, we implement a residual operation on the acquired features in this context: $\hat{\mathbf{R}}^{{\mathcal{P}}/{\mathcal{Q}}} = \hat{\mathbf{H}}^{{\mathcal{P}}/{\mathcal{Q}}} - \hat{\mathbf{F}}^{{\mathcal{P}}/{\mathcal{Q}}}$. Drawing inspiration from the estimator-free registration pipeline \cite{Li2021PointNetLKR} and point cloud classification research \cite{Qi2016PointNetDL}, we concatenate the resulting residual features: $\textbf{Cat}(\hat{\mathbf{R}}^{\mathcal{P}},\hat{\mathbf{R}}^{\mathcal{Q}}) \in \mathbb{R}^{|\mathcal{P} + \mathcal{Q}| \times d}$ and then conduct a max-pooling operation to derive the global feature: $\mathbf{G} \in \mathbb{R}^{1 \times d}$. This diversity furnishes a pivotal cue for the network to grasp and assess the alignment quality of various candidate poses. Subsequently, we employ a three-layer Multilayer Perceptron (MLP) to map and acquire the corresponding confidence scores. After the initial two linear layers, each layer is followed by 1D batch normalization, a leaky ReLU, and a dropout layer. Given that we frame the problem as binary classification, we apply the sigmoid layer subsequent to the final linear layer. The detailed network structure is outlined in the Appendix.

\subsection{Loss Function}

Distinguishing between correct and incorrect poses can be framed as a binary classification task, thus, we can utilize cross-entropy loss to train the network:
\begin{equation}
\mathcal{L}=y \log f_\theta(x)+(1-y) \log \left(1-f_\theta(x)\right)
\label{eq3}
\end{equation}
where $x$ denotes the input pose, $f_{\theta}(x)$ stands for the network output, $\theta$ represents the model parameters, and $y$ represents the ground-truth label, which is either 1 or 0, indicating whether the input pose $x$ is correct or incorrect.

However, directly employing \cref{eq3} would treat each candidate pose equally. Our ultimate objective is to select the optimal transformation from the candidate poses, thereby making it more crucial to learn the evaluation capability of correct poses compared to that of incorrect poses with high errors. Since a distinction exists between the input pose $x$ and the ground truth pose, denoted as $d(x)$, we incorporate a smoothing operation into the loss function:
\begin{equation}
\mathcal{L} = y \cdot (\alpha \cdot (\beta - d(x))) \cdot \log f_\theta(x) + (1 - y) \cdot (1 - \alpha \cdot (d(x) - \beta))^{\gamma} \cdot \log(1 - f_\theta(x))
\end{equation}
where $\alpha$ and $\gamma$ are weighting parameters. Typically, $\beta$ serves as a threshold used to determine whether the pose is correct or incorrect. When the ground truth label $y=1$, $d(x)$ is always less than $\beta$. A smaller $d(x)$ indicates a more accurate pose, hence, $\beta - d(x)$ exerts a greater impact on the loss. When $d(x)$ approaches the boundary, its influence on the loss diminishes, and vice versa. Here, $d(x)$ represents the Root Mean Square Error (RMSE) of the ground-truth correspondences under input pose $x$.

\section{Experiments}
\subsection{Datasets}
We evaluate our model on three public benchmarks, including indoor, outdoor and multi-way registration
scenes. Follow~\cite{huang2021predator}, for indoor scenes, we evaluate our model on both 3DMatch, where
point cloud pairs share > 30\% overlap, and 3DLoMatch, where point cloud pairs have 10\%$\sim$30\% overlap. In line with existing works~\cite{chen2022sc2, chen2023sc}, we evaluate for outdoor scenes on odometry KITTI. In addition, following ~\cite{chen2023sc}, we also use augmentation ICL\_NUIM dataset for testing the performance on the multi-way registration task.  It should be noted that indoor scenes remain the most challenging, particularly on the 3DLoMatch benchmark with lower overlap ratios. Therefore, in the main text, we primarily present the experimental results for this part, while the relevant experiments for outdoor and multi-way registration scenarios are provided in the Appendix.

\subsection{Experimental Setup}
\noindent{\textbf{Metrics.}} 
For indoor scenes, we adopt three typically-used metrics, including Relative Rotation Error (RRE), Relative Translation  Error (RTE) and  Registration Recall (RR). Specifically, RRE and RTE represent the rotation and translation errors between the estimated pose and the ground truth pose, respectively. The RR is defined as the fraction of the point cloud pairs whose RRE and RTE are both below certain thresholds (i.e. RRE < $15^{\circ}$, RTE < 0.3m). In addition, we propose a novel metric, Failure Scenes Recognition Recall (FSRR), to assess the capability of a pose evaluator in identifying the proportion of point cloud pairs in candidate poses that do not contain the correct transformation.

\noindent{\textbf{Implementation details.}} 
To construct the training and validation sets, we utilized the Geotransformer \cite{Qin2022GeometricTF} to extract correspondences for each point cloud pair and employed the SC2-PCR \cite{chen2022sc2} to generate sufficient candidate poses. During the training phase, we ensured that each batch contained 10 correct poses and 10 incorrect poses to maintain a balanced distribution of positive and negative samples. The network is trained with the Adam optimizer for 40 epochs, and the learning rate starts from $1 \times 10^{-6}$ and decays exponentially by 0.05 every epoch. In the feature extraction module, the linear layer maps uniformly to dimension $d=256$. In the pose-aware attention module, the number of nearest neighbors $k$ is set to 16. For the parameters of the loss function, $\alpha$, $\beta$, and $\gamma$ are set to 5, 0.2, and 2, respectively, ensuring that the weight terms are within the range of 0 to 1. In the pose confidence prediction module, the parameter of the dropout layer is set to 0.5. In the actual test, to improve the speed of pose evaluation of our model, CC~\cite{Fischler1981RandomSC} is used to preprocess the candidate poses $\mathcal{H}$, and $|\mathcal{H}|* \delta$ poses are selected in descending order according to their numerical counts. Based on our experience, we have found that setting $\delta=0.4$ and approximately 400 candidate poses achieves a balance between performance and speed. See more implementation details in Appendix.

\begin{table}[h]
\vspace{-10pt}
\caption{Evaluation results with different pose evaluators on 3DMatch and 3DLoMatch.}
\vspace{-0pt}
\setlength{\tabcolsep}{11pt}
\centering
\scriptsize
\renewcommand{\arraystretch}{1.0}
\label{tab.1}
\begin{tabular}{lcccccc} 
\toprule
        & \multicolumn{6}{c}{FPFH}                                                                          \\ 
\cmidrule{2-7}
        & \multicolumn{3}{c}{3DMatch}                    & \multicolumn{3}{c}{3DLoMatch}                    \\ 
\cmidrule(l){2-4} \cmidrule(l){5-7}
        & RR (\%) $\uparrow$       & RRE (deg) $\downarrow$      & RTE (cm) $\downarrow$     & RR (\%) $\uparrow$       & RRE (deg) $\downarrow$    & RTE (cm) $\downarrow$        \\ 
\midrule
CC \cite{Fischler1981RandomSC}      & 83.98          & 2.24          & 6.80          & 36.83          & 4.16          & \uline{10.23}   \\
MAE \cite{Yang2021TowardEA}      & 83.92          & \uline{2.18}  & 6.76          & 37.84          & \uline{3.95}  & 10.08           \\
MSE \cite{Yang2021TowardEA}     & 83.30          & \textbf{2.14} & \uline{6.74}  & 37.11          & \textbf{3.86} & \textbf{9.94}   \\
FS-TCD \cite{chen2023sc}  & \uline{87.18}  & 2.19          & 6.81          & \uline{42.45}  & 4.01          & 10.55           \\
Deep-PE ({\textit{ours}}) & \textbf{89.40} & \uline{2.18}  & \textbf{6.52} & \textbf{50.87} & 4.32          & 11.24           \\ 
\midrule
        & \multicolumn{6}{c}{FCGF}                                                                          \\ 
\cmidrule{2-7}
        & \multicolumn{3}{c}{3DMatch}                    & \multicolumn{3}{c}{3DLoMatch}                    \\ 
\cmidrule(l){2-4} \cmidrule(l){5-7}
        & RR (\%) $\uparrow$        & RRE (deg) $\downarrow$    & RTE (cm) $\downarrow$     & RR (\%) $\uparrow$        & RRE (deg) $\downarrow$    & RTE (cm) $\downarrow$       \\ 
\midrule
CC \cite{Fischler1981RandomSC}      & 94.15          & 1.92          & 6.28          & 57.83          & 3.82          & \uline{10.51}   \\
MAE \cite{Yang2021TowardEA}     & 94.02          & \uline{1.87}  & 6.24          & 58.45          & 3.84          & 10.52           \\
MSE \cite{Yang2021TowardEA}     & 94.09          & \textbf{1.86} & \textbf{6.20} & 58.45          & \uline{3.81}  & \uline{10.51}   \\
FS-TCD \cite{chen2023sc}  & \uline{94.15}  & \uline{1.87}  & \textbf{6.20} & \uline{61.15}  & \textbf{3.68} & \textbf{10.38}  \\
Deep-PE ({\textit{ours}}) & \textbf{95.07} & \uline{1.87}  & 6.24          & \textbf{72.71} & 3.97          & 10.89           \\ 
\midrule
        & \multicolumn{6}{c}{Geotransformer}                                                                \\ 
\cmidrule{2-7}
        & \multicolumn{3}{c}{3DMatch}                    & \multicolumn{3}{c}{3DLoMatch}                    \\ 
\cmidrule(l){2-4} \cmidrule(l){5-7}
        & RR (\%) $\uparrow$       & RRE (deg) $\downarrow$    & RTE (cm) $\downarrow$     & RR (\%) $\uparrow$       & RRE (deg) $\downarrow$    & RTE (cm) $\downarrow$       \\ 
\midrule
CC \cite{Fischler1981RandomSC}      & 95.69          & \textbf{1.92} & 5.75          & 77.99          & 3.01          & 8.66            \\
MAE \cite{Yang2021TowardEA}     & 95.63          & \textbf{1.92} & \uline{5.73}  & 78.33          & 2.97          & 8.60            \\
MSE \cite{Yang2021TowardEA}     & 95.69          & \textbf{1.92} & 5.75          & 78.44          & 2.96          & 8.62            \\
FS-TCD \cite{chen2023sc}  & \uline{95.75}  & \textbf{1.92} & \textbf{5.67} & \uline{78.72}  & \textbf{2.95} & \textbf{8.47}   \\
Deep-PE ({\textit{ours}}) & \textbf{96.24} & \textbf{1.92} & 5.78          & \textbf{80.24} & \textbf{2.95} & \uline{8.52}    \\
\bottomrule
\end{tabular}
\vspace{-6pt}
\end{table}

\subsection{Evaluation on Indoor Scenes}
\noindent{\textbf{Comparison with Different Evaluators. }} 
We compared our method with all known pose evaluators, including Correspondences Counting (CC)~\cite{Fischler1981RandomSC}, Mean Average Error (MAE)~\cite{Yang2021TowardEA}, Mean Square Error (MSE)~\cite{Yang2021TowardEA}, and Feature and Spatial consistency constrained Truncated Chamfer Distance (FS-TCD)~\cite{chen2023sc}. It should be noted that SC2-PCR~\cite{chen2022sc2} is used here to produce candidate poses, and different pose evaluators are used to select the final transformation.
As shown in Tab.~\ref{tab.1}, the improvements in MAE~\cite{Yang2021TowardEA} and MSE~\cite{Yang2021TowardEA} are very limited compared to the most commonly used CC~\cite{Fischler1981RandomSC}, which is also reported in the latest work~\cite{zhang2023MAC}. By establishing more correspondences and combining chamfer distance to analyze the fitting quality of the point clouds, FS-TCD~\cite{chen2023sc} shows some improvements. However, the RR of our method is 2.22\%/8.42\% higher than that of statistics-based pose evaluators when using FPFH~\cite{Rusu2009FastPF}, 0.92\%/11.56\% higher when using FCGF~\cite{Choy2019FullyCG}, and 0.49\%/1.49\% higher when using Geotransformer~\cite{Qin2022GeometricTF}. It's worth noting that RRE/RTE represent the mean errors of successfully registered point cloud pairs. As mentioned in \cite{chen2023sc}, this metric makes methods with high RR more likely to produce larger average errors because they include more challenging data when computing the average error. Nevertheless, Deep-PE still achieves competitive results on RRE and RTE.

\begin{figure*}[h]
  \centering
  \includegraphics[width=\linewidth]{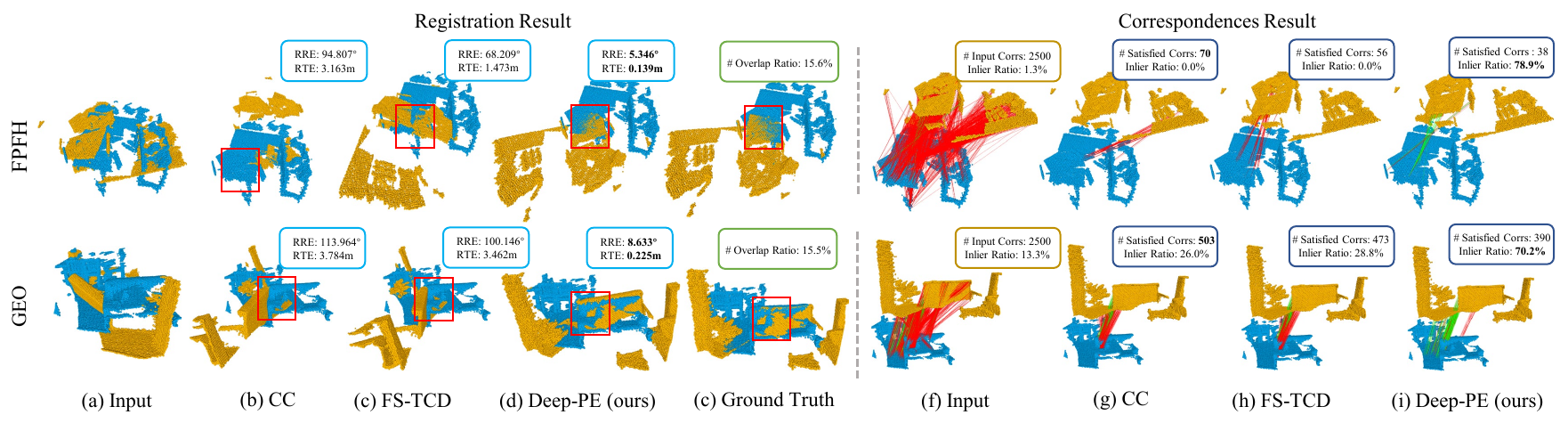}
  \vspace{-15pt}
  \caption{Comparison of registration results between Deep-PE and statistics-based pose evaluators.
  }
  \label{fig.5}
  \vspace{-5pt}
\end{figure*}

In addition, we also present some qualitative results on the 3DLoMatch benchmark. As illustrated in \cref{fig.5}, we use handcrafted FPFH \cite{Rusu2009FastPF} and the advanced Geotransformer (GEO) \cite{Qin2022GeometricTF} extract correspondences respectively, we compared our method to the commonly used CC \cite{Fischler1981RandomSC} and the advanced FS-TCD~\cite{chen2023sc}. We reported the visual registration and correspondence results in challenging scenarios. When the inlier ratio of input correspondences is low, statistics-based  evaluation mechanisms become ineffective, leading to registration failures. In contrast, our method incorporates global alignment priors and utilizes a carefully designed neural network to learn and evaluate the quality of pose alignment. Consequently, it consistently identifies the optimal pose among the candidates, even if it does not satisfy the majority of correspondences.

\noindent{\textbf{Combination with Different Estimators.}} 
We compared our method with commonly used and advanced pose evaluators CC \cite{Fischler1981RandomSC} and FS-TCD \cite{chen2023sc}, incorporating various pose estimators including SM \cite{Leordeanu2005AST}, PointDSC \cite{Bai2021PointDSCRP}, and SC2-PCR \cite{chen2022sc2}. Similarly, each method was tested under FPFH \cite{Rusu2009FastPF} and FCGF \cite{Choy2019FullyCG}. Additionally, for a more vivid comparison of the time efficiency advantages, we also provided registration results and time costs for the classical estimator RANSAC \cite{Fischler1981RandomSC} and the advanced estimator MAC \cite{zhang2023MAC} as references. As shown in \cref{tab.2}, compared to the baseline evaluator CC \cite{Fischler1981RandomSC}, although FS-TCD \cite{chen2023sc} achieved some performance improvements on 3DMatch, its statistics-based evaluation mechanism still overly relies on the quality of input correspondences, resulting in limited improvements on the challenging 3DLoMatch benchmark with lower inlier ratio. In contrast, our method demonstrated significant enhancements in both 3DMatch and 3DLoMatch benchmarks.

In terms of time overhead, as a pioneering learning-based pose evaluator, we have implemented lightweight and batch processing for candidate poses. Nevertheless, our approach still incurs some time cost compared to traditional pose evaluators. However, considering the significant performance improvements achieved, this is worthwhile. Furthermore, the latest MAC \cite{zhang2023MAC} enhances pose estimation accuracy by constructing complex compatibility subgraphs but at a considerable time expense. In contrast, our method achieves notable performance improvements with less time overhead.

\begin{table}[h]
\vspace{-5pt}
\caption{Performance boosting for pose estimators when combined with Deep-PE.}
\centering
\scriptsize
\setlength{\tabcolsep}{6pt}
\renewcommand{\arraystretch}{1.0}
\label{tab.2}
\begin{tabular}{lccccc} 
\toprule
                 & \multicolumn{2}{c}{3DMatch RR(\%)$\uparrow$} & \multicolumn{2}{c}{3DLoMatch RR(\%)$\uparrow$} & Time(s)$\downarrow$  \\ \cmidrule(l){2-3} \cmidrule(l){4-5}
                 & FPFH          & FCGF                         & FPFH          & FCGF                           &                      \\ 
\midrule
RANSAC* \cite{Fischler1981RandomSC}          & 66.10         & 91.44                        & 5.50          & 46.38                          & 2.86                 \\
MAC* \cite{zhang2023MAC}             & 84.10         & 93.72                        & 40.88         & 59.85                          & 3265.42              \\ 
\midrule
SM \cite{Leordeanu2005AST} + CC \cite{Fischler1981RandomSC}            & 55.88         & 86.57                        & 12.34         & 51.43                          & 0.03                 \\
PointDSC \cite{Bai2021PointDSCRP} + CC \cite{Fischler1981RandomSC}      & 77.57         & 92.85                        & 20.38         & 56.09                          & 0.10                 \\
SC2-PCR \cite{chen2022sc2} + CC \cite{Fischler1981RandomSC}       & 83.98         & 93.28                        & 38.57         & 57.83                          & 0.11                 \\ 
\midrule
SM \cite{Leordeanu2005AST} + FS-TCD \cite{chen2023sc}        & 60.75 (+4.87) & 87.00 (+0.53)                & 14.74 (+2.44) & 51.21 (+2.78)                  & 0.18 (+0.15)         \\
PointDSC \cite{Bai2021PointDSCRP} + FS-TCD \cite{chen2023sc}  & 82.62 (+5.05) & 93.47 (+0.62)                & 23.09 (+2.71) & 59.23 (+3.14)                  & 0.25 (+0.15)         \\
SC2-PCR \cite{chen2022sc2} + FS-TCD \cite{chen2023sc}   & 87.18 (+3.20) & 94.15 (+0.87)                & 41.47 (+2.90) & 61.15 (+3.32)                  & 0.26 (+0.15)         \\ 
\midrule
SM \cite{Leordeanu2005AST} + Deep-PE ({\textit{ours}})       & 63.52 (+7.64) & 88.46 (+1.89)                & 28.22 (+15.88)& 63.55 (+12.12)                 & 0.64 (+0.61)         \\
PointDSC \cite{Bai2021PointDSCRP} + Deep-PE ({\textit{ours}}) & 85.08 (+7.51) & 94.21 (+1.36)                & 34.72 (+14.34)& 67.84 (+11.75)                 & 0.71 (+0.61)         \\
SC2-PCR \cite{chen2022sc2} + Deep-PE ({\textit{ours}})  & 89.40 (+5.42) & 94.82 (+1.54)                & 50.25 (+12.19)& 71.87 (+14.04)                 & 0.72 (+0.61)         \\
\bottomrule
\end{tabular}
\label{tab.2}
\vspace{-23pt}
\end{table}

\noindent{\textbf{Robustness to Low Inlier Ratio.}} 
 We extracted correspondences using FPFH \cite{Rusu2009FastPF} separately on the 3DMatch and 3DLoMatch datasets, followed by generating candidate poses using SC2-PCR \cite{chen2022sc2}. Subsequently, based on the inlier ratio, we categorized all point cloud pairs into six different groups. As illustrated in   \cref{fig.6} and \cref{fig1b}, when the inlier ratio drops below 1\%, the RR of statistics-based evaluators \cite{Fischler1981RandomSC, chen2023sc} is extremely low, while Deep-PE demonstrates commendable performance, highlighting the robustness of our method in scenarios with low inlier ratios.

\begin{minipage}[t]{\textwidth}
    \begin{minipage}[t]{0.5\textwidth}
    \centering
    \includegraphics[width=1.0\linewidth]{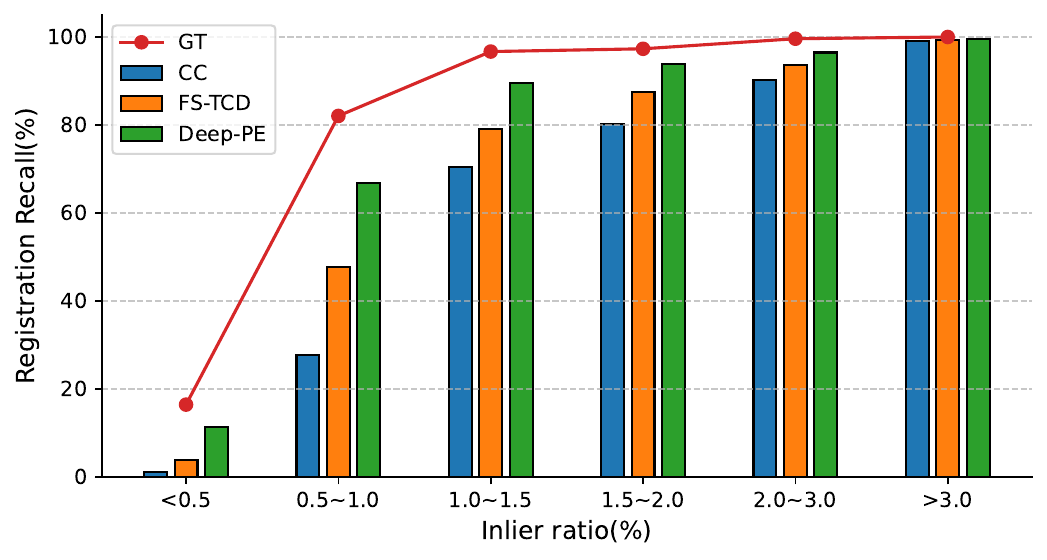}
    \vspace{-15pt}
    \captionof{figure}{The registration recall under different inlier ratios on 3DMatch.}
    \label{fig.6}
    \end{minipage}
    \hfill
    \begin{minipage}[t]{0.5\textwidth}
    \centering
    \vspace{-100pt}
    \captionof{table}{Failure scenes recognition recall with different $\lambda$.}
    \vspace{-5pt}
    \scriptsize
    \setlength{\tabcolsep}{5.5pt}
    \begin{tabular}{ccccccc}
    \toprule
                  & \multicolumn{3}{c}{3DMatch FSRR(\%)$\uparrow$} & \multicolumn{3}{c}{3DLoMatch FSRR(\%)$\uparrow$} 
                  \\ \cmidrule(l){2-4} \cmidrule(l){5-7}
                  & FPFH           & FCGF          & GEO           & FPFH           & FCGF           & GEO            \\ \midrule
    $\lambda$=0.1 & 44.92          & 4.00          & 25.00         & 21.38          & 9.18           & 10.44          \\
    $\lambda$=0.2 & 54.55          & 12.00         & 27.78         & 32.51          & 14.12          & 12.46          \\
    $\lambda$=0.3 & 60.96          & 12.00         & 41.67         & 40.67          & 19.53          & 16.84          \\
    $\lambda$=0.4 & 63.10          & 20.00         & 41.67         & 47.71          & 26.59          & 21.21          \\
    $\lambda$=0.5 & 69.52          & 22.00         & 44.44         & 52.41          & 32.71          & 26.60          \\
    $\lambda$=0.6 & 73.80          & 26.00         & 50.00         & 55.62          & 39.06          & 32.32          \\
    $\lambda$=0.7 & 77.01          & 32.00         & 50.00         & 62.18          & 44.94          & 37.37          \\
    $\lambda$=0.8 & 79.68          & 32.00         & 50.00         & 68.23          & 50.82          & 45.12          \\
    $\lambda$=0.9 & 84.49          & 48.00         & 58.33         & 76.39          & 60.47          & 54.88          \\ \bottomrule
    \end{tabular}
    \label{tab.6}
    \end{minipage}
\end{minipage}


\noindent{\textbf{Failure Scenes Recognition Capability Analysis.}} 
 Compared to traditional statistics-based pose evaluators, our method can identify point cloud pairs within candidate poses lacking correct transformations. We employ SC2-PCR \cite{chen2022sc2} as the pose estimator and then utilize CC \cite{Fischler1981RandomSC} to pre-select the top 100 most challenging candidate poses for each point cloud pair. In the 3DMatch/3DLoMatch benchmarks, the numbers of failure point cloud pairs generated by FPFH \cite{Rusu2009FastPF}, FCGF \cite{Choy2019FullyCG}, and Geotransformer (GEO) \cite{Qin2022GeometricTF} are 187/809, 50/425, and 36/297, respectively. As we treat this issue as a binary classification task, the confidence threshold $\lambda$ can vary within the range of [0,1] to analyze our method's ability to identify registration failure scenes. As shown in \cref{tab.3}, with the increase of the confidence threshold, FSRR also increases correspondingly. This indicates that our predicted confidence scores effectively reflect the likelihood of successful registration. Compared to traditional pose evaluators \cite{Fischler1981RandomSC,Yang2021TowardEA, chen2023sc}, our method avoids the dilemma of selecting from candidate poses lacking correct transformations, which holds significant practical implications.

\section{Conclusion}


In this paper, we introduce Deep-PE as a pioneering approach that leverages deep learning to determine the optimal pose without necessitating explicit initial correspondences. The architecture of Deep-PE is meticulously crafted and lightweight, featuring core components including a pose-aware attention module that simulates and learns alignment status under various candidate poses, and a pose confidence prediction module that predicts the confidence for each candidate pose, ultimately selecting the pose with the highest confidence score as the final transformation. Moreover, we propose a weighted cross-entropy loss to re-weight the loss of candidate poses based on their similarity to the ground truth, thereby enhancing the capability to select the optimal pose. Our experiments across multiple public benchmarks demonstrate that Deep-PE achieves state-of-the-art results and showcases commendable performance in low-overlap scenarios with poor input correspondence quality, such as those encountered in 3DLoMatch.

\bibliographystyle{plain}
\bibliography{ref}

\begin{thebibliography}{10}

\bibitem{Ao2020SpinNetLA}
Sheng Ao, Qingyong Hu, Bo~Yang, A.~Markham, and Yulan Guo.
\newblock Spinnet: Learning a general surface descriptor for 3d point cloud registration.
\newblock {\em 2021 IEEE/CVF Conference on Computer Vision and Pattern Recognition (CVPR)}, pages 11748--11757, 2020.

\bibitem{Bai2021PointDSCRP}
Xuyang Bai, Zixin Luo, Lei Zhou, Hongkai Chen, Lei Li, Zeyu Hu, Hongbo Fu, and Chiew-Lan Tai.
\newblock Pointdsc: Robust point cloud registration using deep spatial consistency.
\newblock {\em 2021 IEEE/CVF Conference on Computer Vision and Pattern Recognition (CVPR)}, pages 15854--15864, 2021.

\bibitem{Bai2020D3FeatJL}
Xuyang Bai, Zixin Luo, Lei Zhou, Hongbo Fu, Long Quan, and Chiew-Lan Tai.
\newblock D3feat: Joint learning of dense detection and description of 3d local features.
\newblock {\em 2020 IEEE/CVF Conference on Computer Vision and Pattern Recognition (CVPR)}, pages 6358--6366, 2020.

\bibitem{Barth2017GraphCutR}
D{\'a}niel Bar{\'a}th and Jiri Matas.
\newblock Graph-cut ransac.
\newblock {\em 2018 IEEE/CVF Conference on Computer Vision and Pattern Recognition}, pages 6733--6741, 2017.

\bibitem{Besl1992AMF}
Paul~J. Besl and Neil~D. McKay.
\newblock A method for registration of 3-d shapes.
\newblock {\em IEEE Trans. Pattern Anal. Mach. Intell.}, 14:239--256, 1992.

\bibitem{chen2023sc}
Zhi Chen, Kun Sun, Fan Yang, Lin Guo, and Wenbing Tao.
\newblock Sc2-pcr++: Rethinking the generation and selection for efficient and robust point cloud registration.
\newblock {\em IEEE Transactions on Pattern Analysis and Machine Intelligence}, 2023.

\bibitem{chen2022sc2}
Zhi Chen, Kun Sun, Fan Yang, and Wenbing Tao.
\newblock Sc2-pcr: A second order spatial compatibility for efficient and robust point cloud registration.
\newblock In {\em Proceedings of the IEEE/CVF Conference on Computer Vision and Pattern Recognition}, pages 13221--13231, 2022.

\bibitem{Choy2020DeepGR}
Christopher~Bongsoo Choy, Wei Dong, and Vladlen Koltun.
\newblock Deep global registration.
\newblock {\em 2020 IEEE/CVF Conference on Computer Vision and Pattern Recognition (CVPR)}, pages 2511--2520, 2020.

\bibitem{Choy2019FullyCG}
Christopher~Bongsoo Choy, Jaesik Park, and Vladlen Koltun.
\newblock Fully convolutional geometric features.
\newblock {\em 2019 IEEE/CVF International Conference on Computer Vision (ICCV)}, pages 8957--8965, 2019.

\bibitem{Fischler1981RandomSC}
Martin~A. Fischler and Robert~C. Bolles.
\newblock Random sample consensus: a paradigm for model fitting with applications to image analysis and automated cartography.
\newblock {\em Commun. ACM}, 24:381--395, 1981.

\bibitem{Geiger2012AreWR}
Andreas Geiger, Philip Lenz, and Raquel Urtasun.
\newblock Are we ready for autonomous driving? the kitti vision benchmark suite.
\newblock {\em 2012 IEEE Conference on Computer Vision and Pattern Recognition}, pages 3354--3361, 2012.

\bibitem{Gojcic2018ThePM}
Zan Gojcic, Caifa Zhou, Jan~Dirk Wegner, and Andreas Wieser.
\newblock The perfect match: 3d point cloud matching with smoothed densities.
\newblock {\em 2019 IEEE/CVF Conference on Computer Vision and Pattern Recognition (CVPR)}, pages 5540--5549, 2018.

\bibitem{Handa2014ABF}
Ankur Handa, Thomas Whelan, John~B. McDonald, and Andrew~J. Davison.
\newblock A benchmark for rgb-d visual odometry, 3d reconstruction and slam.
\newblock {\em 2014 IEEE International Conference on Robotics and Automation (ICRA)}, pages 1524--1531, 2014.

\bibitem{he2020epipolar}
Yihui He, Rui Yan, Katerina Fragkiadaki, and Shoou-I Yu.
\newblock Epipolar transformers.
\newblock In {\em Proceedings of the ieee/cvf conference on computer vision and pattern recognition}, pages 7779--7788, 2020.

\bibitem{huang2021predator}
Shengyu Huang, Zan Gojcic, Mikhail Usvyatsov, Andreas Wieser, and Konrad Schindler.
\newblock Predator: Registration of 3d point clouds with low overlap.
\newblock In {\em Proceedings of the IEEE/CVF Conference on computer vision and pattern recognition}, pages 4267--4276, 2021.

\bibitem{Huang2020FeatureMetricRA}
Xiaoshui Huang, Guofeng Mei, and Jian Zhang.
\newblock Feature-metric registration: A fast semi-supervised approach for robust point cloud registration without correspondences.
\newblock {\em 2020 IEEE/CVF Conference on Computer Vision and Pattern Recognition (CVPR)}, pages 11363--11371, 2020.

\bibitem{jian2010robust}
Bing Jian and Baba~C Vemuri.
\newblock Robust point set registration using gaussian mixture models.
\newblock {\em IEEE transactions on pattern analysis and machine intelligence}, 33(8):1633--1645, 2010.

\bibitem{jiang2023robust}
Haobo Jiang, Zheng Dang, Zhen Wei, Jin Xie, Jian Yang, and Mathieu Salzmann.
\newblock Robust outlier rejection for 3d registration with variational bayes.
\newblock In {\em Proceedings of the IEEE/CVF conference on computer vision and pattern recognition}, pages 1148--1157, 2023.

\bibitem{Leordeanu2005AST}
Marius Leo and Martial Hebert.
\newblock A spectral technique for correspondence problems using pairwise constraints.
\newblock {\em Tenth IEEE International Conference on Computer Vision (ICCV'05) Volume 1}, 2:1482--1489 Vol. 2, 2005.

\bibitem{Li2021PointNetLKR}
Xueqian Li, Jhony~Kaesemodel Pontes, and Simon Lucey.
\newblock Pointnetlk revisited.
\newblock {\em 2021 IEEE/CVF Conference on Computer Vision and Pattern Recognition (CVPR)}, pages 12758--12767, 2021.

\bibitem{lu2019deepvcp}
Weixin Lu, Guowei Wan, Yao Zhou, Xiangyu Fu, Pengfei Yuan, and Shiyu Song.
\newblock Deepvcp: An end-to-end deep neural network for point cloud registration.
\newblock In {\em Proceedings of the IEEE/CVF international conference on computer vision}, pages 12--21, 2019.

\bibitem{mei2023unsupervised}
Guofeng Mei, Hao Tang, Xiaoshui Huang, Weijie Wang, Juan Liu, Jian Zhang, Luc Van~Gool, and Qiang Wu.
\newblock Unsupervised deep probabilistic approach for partial point cloud registration.
\newblock In {\em Proceedings of the IEEE/CVF Conference on Computer Vision and Pattern Recognition}, pages 13611--13620, 2023.

\bibitem{Myatt2002NAPSACHN}
Darren~R. Myatt, Philip H.~S. Torr, Slawomir~Jaroslaw Nasuto, J.~Mark Bishop, and R.~Craddock.
\newblock Napsac: High noise, high dimensional robust estimation - it's in the bag.
\newblock In {\em British Machine Vision Conference}, 2002.

\bibitem{Pais20193DRegNetAD}
G.~Dias Pais, Pedro Miraldo, Srikumar Ramalingam, Venu~Madhav Govindu, Jacinto~C. Nascimento, and Rama Chellappa.
\newblock 3dregnet: A deep neural network for 3d point registration.
\newblock {\em 2020 IEEE/CVF Conference on Computer Vision and Pattern Recognition (CVPR)}, pages 7191--7201, 2019.

\bibitem{Papadopoulo2000EstimatingTJ}
Th{\'e}odore Papadopoulo and Manolis I.~A. Lourakis.
\newblock Estimating the jacobian of the singular value decomposition: Theory and applications.
\newblock In {\em European Conference on Computer Vision}, 2000.

\bibitem{Qi2016PointNetDL}
C.~Qi, Hao Su, Kaichun Mo, and Leonidas~J. Guibas.
\newblock Pointnet: Deep learning on point sets for 3d classification and segmentation.
\newblock {\em 2017 IEEE Conference on Computer Vision and Pattern Recognition (CVPR)}, pages 77--85, 2016.

\bibitem{Qin2022GeometricTF}
Zheng Qin, Hao Yu, Changjian Wang, Yulan Guo, Yuxing Peng, and Kaiping Xu.
\newblock Geometric transformer for fast and robust point cloud registration.
\newblock {\em 2022 IEEE/CVF Conference on Computer Vision and Pattern Recognition (CVPR)}, pages 11133--11142, 2022.

\bibitem{Rusu2009FastPF}
Radu~Bogdan Rusu, Nico Blodow, and Michael Beetz.
\newblock Fast point feature histograms (fpfh) for 3d registration.
\newblock {\em 2009 IEEE International Conference on Robotics and Automation}, pages 3212--3217, 2009.

\bibitem{Sun2021LoFTRDL}
Jiaming Sun, Zehong Shen, Yuang Wang, Hujun Bao, and Xiaowei Zhou.
\newblock Loftr: Detector-free local feature matching with transformers.
\newblock {\em 2021 IEEE/CVF Conference on Computer Vision and Pattern Recognition (CVPR)}, pages 8918--8927, 2021.

\bibitem{Thomas2019KPConvFA}
Hugues Thomas, C.~Qi, Jean-Emmanuel Deschaud, Beatriz Marcotegui, François Goulette, and Leonidas~J. Guibas.
\newblock Kpconv: Flexible and deformable convolution for point clouds.
\newblock {\em 2019 IEEE/CVF International Conference on Computer Vision (ICCV)}, pages 6410--6419, 2019.

\bibitem{Tombari2010UniqueSO}
Federico Tombari, Samuele Salti, and Luigi di~Stefano.
\newblock Unique signatures of histograms for local surface description.
\newblock In {\em European Conference on Computer Vision}, 2010.

\bibitem{Vaswani2017AttentionIA}
Ashish Vaswani, Noam~M. Shazeer, Niki Parmar, Jakob Uszkoreit, Llion Jones, Aidan~N. Gomez, Lukasz Kaiser, and Illia Polosukhin.
\newblock Attention is all you need.
\newblock {\em ArXiv}, abs/1706.03762, 2017.

\bibitem{Wang2019SuperGLUEAS}
Alex Wang, Yada Pruksachatkun, Nikita Nangia, Amanpreet Singh, Julian Michael, Felix Hill, Omer Levy, and Samuel~R. Bowman.
\newblock Superglue: A stickier benchmark for general-purpose language understanding systems.
\newblock {\em ArXiv}, abs/1905.00537, 2019.

\bibitem{Wang2023RoRegPP}
Haiping Wang, Yuan Liu, Qingyong Hu, Bing Wang, Jianguo Chen, Zhen Dong, Yulan Guo, Wenping Wang, and Bisheng Yang.
\newblock Roreg: Pairwise point cloud registration with oriented descriptors and local rotations.
\newblock {\em IEEE transactions on pattern analysis and machine intelligence}, PP, 2023.

\bibitem{Wang2019DeepCP}
Yue Wang and Justin~M. Solomon.
\newblock Deep closest point: Learning representations for point cloud registration.
\newblock {\em 2019 IEEE/CVF International Conference on Computer Vision (ICCV)}, pages 3522--3531, 2019.

\bibitem{wang2019dynamic}
Yue Wang, Yongbin Sun, Ziwei Liu, Sanjay~E Sarma, Michael~M Bronstein, and Justin~M Solomon.
\newblock Dynamic graph cnn for learning on point clouds.
\newblock {\em ACM Transactions on Graphics (tog)}, 38(5):1--12, 2019.

\bibitem{Yang2021TowardEA}
Jiaqi Yang, Zhiqiang Huang, Siwen Quan, Qian Zhang, Yanning Zhang, and Zhiguo Cao.
\newblock Toward efficient and robust metrics for ransac hypotheses and 3d rigid registration.
\newblock {\em IEEE Transactions on Circuits and Systems for Video Technology}, 32:893--906, 2021.

\bibitem{Yew2020RPMNetRP}
Zi~Jian Yew and Gim~Hee Lee.
\newblock Rpm-net: Robust point matching using learned features.
\newblock {\em 2020 IEEE/CVF Conference on Computer Vision and Pattern Recognition (CVPR)}, pages 11821--11830, 2020.

\bibitem{Yu2021CoFiNetRC}
Hao Yu, Fu~Li, Mahdi Saleh, Benjamin Busam, and Slobodan Ilic.
\newblock Cofinet: Reliable coarse-to-fine correspondences for robust point cloud registration.
\newblock In {\em Neural Information Processing Systems}, 2021.

\bibitem{Yuan2020DeepGMRLL}
Wentao Yuan, Benjamin Eckart, Kihwan Kim, V.~Jampani, Dieter Fox, and Jan Kautz.
\newblock Deepgmr: Learning latent gaussian mixture models for registration.
\newblock In {\em European Conference on Computer Vision}, 2020.

\bibitem{Zeng20163DMatchLL}
Andy Zeng, Shuran Song, Matthias Nie{\ss}ner, Matthew Fisher, Jianxiong Xiao, and Thomas~A. Funkhouser.
\newblock 3dmatch: Learning local geometric descriptors from rgb-d reconstructions.
\newblock {\em 2017 IEEE Conference on Computer Vision and Pattern Recognition (CVPR)}, pages 199--208, 2016.

\bibitem{zhang2023MAC}
Xiyu Zhang, Jiaqi Yang, Shikun Zhang, and Yanning Zhang.
\newblock 3d registration with maximal cliques.
\newblock In {\em Proceedings of the IEEE/CVF Conference on Computer Vision and Pattern Recognition}, pages 17745--17754, 2023.

\end{thebibliography}

\renewcommand{\thetable}{{\Alph{table}}}
\renewcommand{\thefigure}{{\Alph{figure}}}
\renewcommand{\thesection}{{\Alph{section}}}
\setcounter{figure}{0}
\setcounter{table}{0}
\setcounter{section}{0}

\clearpage
\section{Visualization of Deep-PE in the Registration Pipeline}

\begin{figure}[h]
\includegraphics[width=\textwidth]{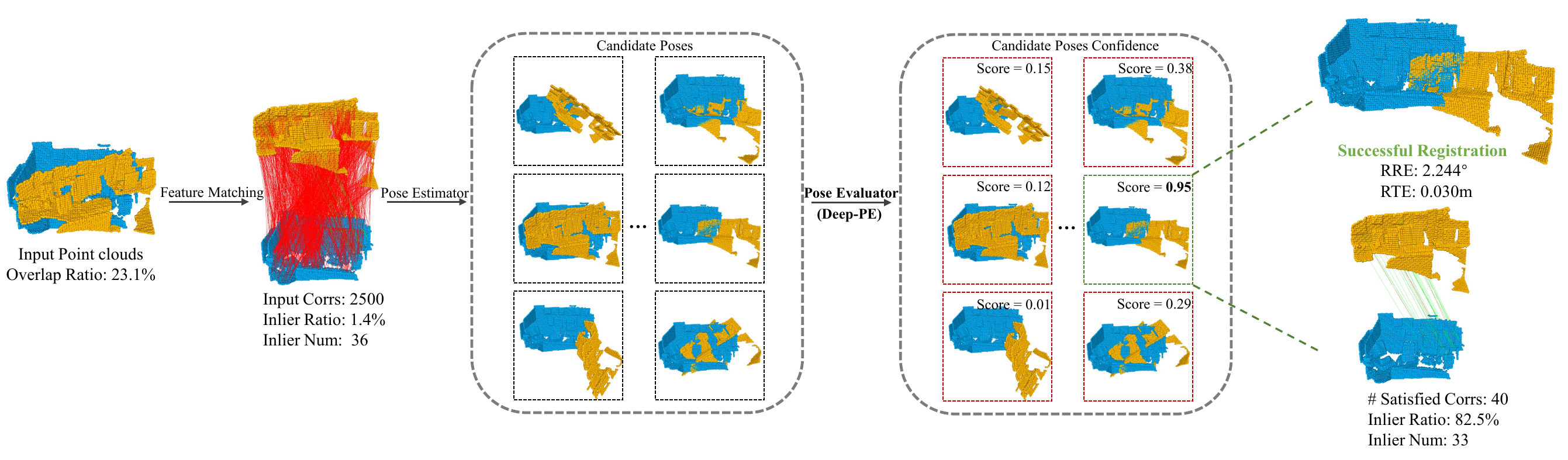}
  \vspace{-12pt}
  \caption{
  Visualization of Deep-PE in the estimator-based registration pipeline. Such pipelines typically involve establishing correspondences between point clouds using feature descriptors and generating a set of candidate poses using a pose estimator. Subsequently, our proposed Deep-PE predicts the confidence for each candidate pose and selects the pose with the highest confidence score as the final transformation.
  RRE: Relative Rotation Error. RTE: Relative Translation Error.
  \#~Input Corrs: No. of input correspondences.
  \#~Satisfied Corrs: No. of correspondences consistent with the predicted pose.
    }
  \label{fig.A}
\end{figure}

Pose evaluation represents the final step in the estimator-based registration pipeline, and despite its paramount importance, it has been significantly overlooked in current research. In order to better illustrate our work, we have undertaken a visualization of each step in this pipeline. As depicted in \cref{fig.A}, for the input point cloud pairs, an initial step involves feature description and matching to generate a set of correspondences. Subsequently, the pose estimator generates a set of candidate poses. Following this, our proposed Deep-PE predicts the
likelihood of successful registration for each candidate pose and selects the pose with the highest score as the final transformation. It is noticeable that due to the influence of the low overlap rate of point clouds, the Inlier Ratio (IR) of extracted correspondences is usually low. In this particular example, the IR is 1.4\%. Fortunately, contemporary robust pose estimators are capable of generating candidate poses that encapsulate the correct solution. Differing from existing methods which excessively depend on the quality of input correspondences, our approach simulates and learns the alignment status of point clouds under various candidate poses, thereby enabling the accurate selection of the correct transformation from candidate poses.

\section{More Insights into Pose Evaluation}
In this section, we provide insights into our focus on the pose evaluation stage and outline the essential criteria for an effective pose evaluator.

\subsection{Pose Evaluator as the Bottleneck}
A typical estimator-based registration pipeline consists of three fundamental steps: feature description and matching, pose estimation, and pose evaluation. In the subsequent sections, we aim to identify the bottleneck step.

\textit{Experimental setting.}
We selected Geotransformer~\cite{Qin2022GeometricTF} for correspondence extraction and SC2-PCR~\cite{chen2022sc2} as the pose estimator to generate a set of candidate poses. Our experiments were performed on the 3DMatch and 3DLoMatch datasets~\cite{Zeng20163DMatchLL}. Using Geotransformer~\cite{Qin2022GeometricTF} and SC2-PCR~\cite{chen2022sc2}, we generated 5000 correspondences and produced 1000 candidate poses. We then applied the widely used Correspondence Counting (CC) as evaluator~\cite{Fischler1981RandomSC} to select the final pose.

\textit{Indicators.}
We utilize the following indicators to analyze registration failures:
\begin{itemize}
    \item \textbf{Total:} This indicator represents the overall recall of registration failures.

    \item \textbf{Des \& Match:} This indicator represents the percentage of cases where deteriorated descriptor performance results in too few correct correspondences (less than three), emphasizing that at least three correct correspondences are essential for accurate pose estimation.

    \item \textbf{Estimator:} This indicator quantifies the proportion of instances where, despite having a minimum of three correct correspondences, the candidate poses generated by the pose estimator fail to include a least one correct transformation.

    \item \textbf{Evaluator:} This indicator shows the percentage of cases where the correct transformations, although present among the candidate poses, is incorrectly overlooked by the pose evaluator in favor of an incorrect final pose selection.
\end{itemize}

\begin{figure}[h]
  \centering
  \includegraphics[width=0.85\linewidth]{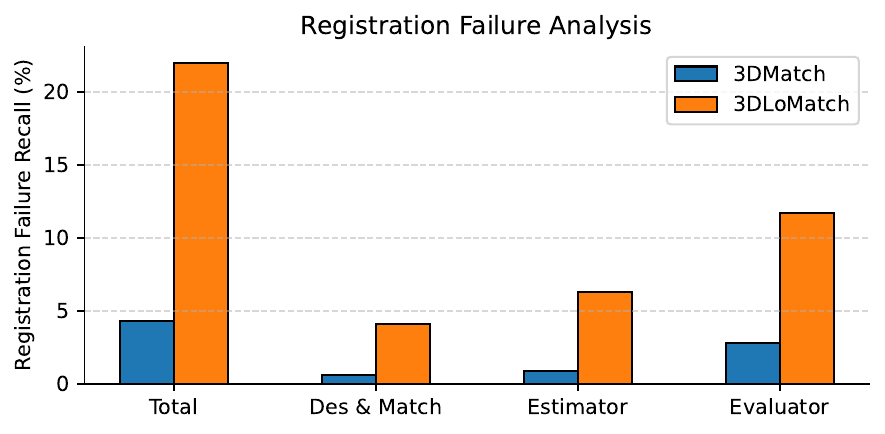}
  \vspace{-7pt}
  \caption{
  Utilizing 3DMatch and 3DLoMatch datasets, we plot the registration failure recall across three stages: Des $\&$ Match (correspondences generation), Estimator (candidate poses estimation), and Evaluator (optimal pose selection). 
  }
  \label{fig.B}
\end{figure}

\textit{Conclusion.}
Analyzing the statistics of registration failure recall for each step, as illustrated in \cref{fig.B}, reveals a significant finding: up to 12\% of failures can be attributed to the pose evaluator, while failures due to feature descriptors (Des \& Match) are limited to approximately 4\%. This observation highlights that the primary bottleneck in the current estimator-based registration pipeline lies with the statistics-based pose evaluators, which impede correct pose selection.

\subsection{Performance Degradation of Statistics-based Evaluators in Low Overlap}

\begin{figure}[h]
  \centering
  \includegraphics[width=0.85\linewidth]{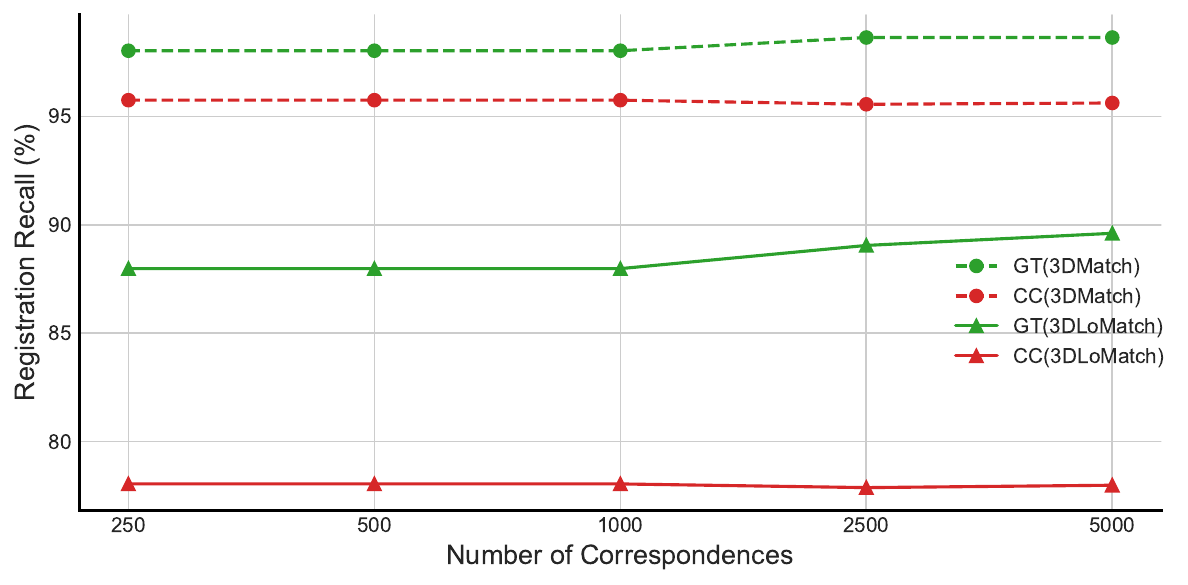}
  \vspace{-8pt}
  \caption{
    The performance of Correspondence Counting (CC) significantly declines on 3DLoMatch, falling well below the ground-truth (GT) performance. Here, GT is defined by the likelihood that at least one correct transformation appears within the candidate poses.
    }
    \label{fig.C}
\end{figure}

In this part, we come to examine the scalability of registration performance for statistics-based pose evaluators in relation to varying numbers of correspondences. The experimental setup remained consistent with the previous subsection, with the distinction of employing the Geotransformer \cite{Qin2022GeometricTF} to generate different quantities of correspondences. Results from experiments on the 3DMatch and 3DLoMatch benchmark datasets are depicted in \cref{fig.C}. On one side, the likelihood of Correspondence Counting (CC) \cite{Fischler1981RandomSC} failing increases in scenarios of low overlap. As shown in \cref{tab.A}, this is particularly evident when the overlap region is either texture-deficient or has a low overlap rate, leading to correspondences that are significantly marred by erroneous outliers. On the opposite spectrum, an increase in the number of correspondences does not necessarily translate to an improved registration recall. In fact, a reduction in the registration recall rate was observed when the sampling numbers reached 2500 and 5000, subsequently impairing the efficacy of CC \cite{Fischler1981RandomSC} in identifying the correct transformation. We also observe that with an increase in the number of correspondences samples, the ground truth (GT) values correspondingly increase. This indicates that the pose estimator can increase the probability of including the correct transformations in the candidate poses. However, the registration recall decreases due to the statistical-based pose evaluator, highlighting the necessity of proposing an evaluator independent of the quality of input correspondences.

\begin{table}[h]
    \caption{Inlier ratio across different correspondence sample numbers.}
    \vspace{0pt}
    \setlength{\tabcolsep}{8pt} 
    \scriptsize
    \centering
    \label{tab.A}
    \begin{tabular}{lcccccccccc} 
    \toprule
                    & \multicolumn{5}{c}{3DMatch IR (\%)$\uparrow$} & \multicolumn{5}{c}{3DLoMatch IR (\%)$\uparrow$}  \\ 
    \cmidrule(l){2-6} \cmidrule(l){7-11}
    \# Samples      & 250  & 500  & 1000 & 2500 & 5000              & 250  & 500  & 1000 & 2500 & 5000                 \\ 
    \cmidrule(l){1-6} \cmidrule(l){7-11}
    Geotransformer \cite{Qin2022GeometricTF} & 85.1 & 82.2 & 76.0 & 75.2 & 71.9              & 57.7 & 52.9 & 46.2 & 45.3 & 43.5                 \\
    \bottomrule
    \end{tabular}
\end{table}

\subsection{Essential Criteria for an Effective Pose Evaluator}
Based on the above discussion, we give essential criteria for an effective pose evaluator.

\textbf{Lightweight Network Design.}
Typically, traditional pose estimation algorithms like RANSAC~\cite{Fischler1981RandomSC} tend to generate a large pool of candidate poses. While some robust estimators, such as SC2-PCR~\cite{chen2022sc2}, are designed to yield a smaller, more precise set of candidate poses, the overall number of these poses remains considerable. Consequently, the development of a versatile, learning-based pose evaluator becomes critical for the efficient evaluation of each candidate pose. This calls for a lightweight network architecture that not only processes candidate poses efficiently but also supports batch processing for enhanced throughput.

\textbf{Perception of Global Alignment.}
The network is tasked with learning to accurately differentiate between correct and incorrect poses by incorporating global alignment information. This differentiation is achievable through the use of statistics-based pose estimators, which analyze the overall spatial relationships and alignment patterns across the entire scene. By focusing on these global characteristics, the network can more effectively identify the validity of each pose, distinguishing those that align well with the global context from those that do not.

\textbf{Pose Sensitivity.}
The network needs to demonstrate sensitivity to variations in poses, requiring it not only to distinguish accurately between correct and incorrect pose hypotheses but also to assess the degrees of accuracy among candidate poses. Specifically, the network should assign a higher confidence score to candidate poses that are exact in their accuracy compared to correct ones that present a degree of deviation. Implementing this capability is crucial for improving the overall effectiveness of pose evaluation, ensuring that the most accurate poses are prioritized and selected.

\textbf{Quality Independent of Input Correspondences.}
The design of the network should prioritize sensitivity to pose variations above all, especially given that a low inlier count among input correspondences can lead to situations where, despite outliers being incorrect, they still demonstrate compatibility and advocate for the same transformation. In these scenarios, statistics-based strategies might not be effective. As depicted in \cref{fig.B}, even robust pose estimators are capable of generating at least one correct pose under such conditions. Consequently, our methodology primarily focuses on evaluating the candidate poses themselves, rather than the quality of the input correspondences. This approach ensures that our proposed network maintains strong robustness across a wide range of conditions, effectively handling the intricacies and challenges posed by various scenarios.

\textbf{Capability to Identify Incorrect poses}
Traditional pose evaluators primarily rely on statistical mechanisms to select the optimal transformation from a set of candidate poses, but they have limitations in recognizing incorrect poses. For instance, in scenarios where the candidate poses lacks the presence of correct transformations, these evaluators may have to choose an incorrect pose as the final transformation. The significance of a versatile pose evaluator lies in its ability to identify incorrect poses, which holds practical value in real-world scenarios. This not only enhances the reliability of the pose estimation process but also substantially reduces the risk of propagating errors, especially in applications where accurate pose estimation is crucial.

\section{Evaluation Metrics}
Following common practice~\cite{huang2021predator, Qin2022GeometricTF}, we use different evaluation metrics to evaluate the quality of correspondences and registration results.

{\noindent \textbf{Inlier Ratio}} (IR). It measures the ratio of putative correspondences whose residual distance is smaller than a threshold ({i.e.} $\tau_1=0.1/0.6$m for indoor/outdoor scenes) under the ground-truth transformation $\bar{\mathbf{T}}_{\mathbf{P}\rightarrow \mathbf{Q}}$:
\begin{equation}
\mathbf{IR}=\frac{1}{|\mathcal{C}|} \sum_{\left(\mathbf{p}_{x_i}, \mathbf{q}_{y_i}\right) \in \mathcal{C}} \llbracket\left\|\bar{\mathbf{T}}_{\mathbf{P} \rightarrow \mathbf{Q}}\left(\mathbf{p}_{x_i}\right)-\mathbf{q}_{y_i}\right\|_2<\tau_1 \rrbracket
\end{equation}
where $\llbracket\cdot\rrbracket$ is the Iverson bracket.

\noindent{\textbf{Relative Rotation Error}} (RRE). It measures the geodesic distance in degrees between estimated and ground-truth rotation matrices.
\begin{equation}
\mathbf{RRE}=\arccos \left(\frac{\operatorname{trace}\left(\mathbf{R}^T \cdot \bar{\mathbf{R}}-1\right)}{2}\right)
\end{equation}

{\noindent \textbf{Relative Translation Error}} (RTE). It measures the Euclidean distance
between estimated and ground-truth translation vectors.
\begin{equation}
\mathbf{RTE}=\|\mathbf{t}-\hat{\mathbf{t}}\|_2
\end{equation}

{\noindent \textbf{Registration Recall}} (RR). It is defined as the fraction of the point cloud pairs whose RRE and RTE are both below certain thresholds:
\begin{equation}
\mathbf{RR}=\frac{1}{M} \sum_{i=1}^M \llbracket \mathbf{RRE}_i<\tau_2 \wedge \mathbf{RTE}_i < \tau_3 \rrbracket
\end{equation}
($\tau_2,\tau_3$) is set to ($15^{\circ}, 0.3$m) and ($5^{\circ}, 0.6$m) for the indoor and outdoor scenes, respectively.

{\noindent \textbf{Failure Scenes Recognition Recall}} (FSRR). We propose a novel metric to assess the capability of a pose evaluator in identifying the proportion of point cloud pairs in candidate poses that do not contain the correct transformation.
\begin{equation}
\mathbf{FSRR}=\frac{1}{N} \sum_{i=1}^N \llbracket \sum_{j=1}^K \llbracket S_j < \lambda  \rrbracket = K \rrbracket
\end{equation}
where $N$ represents the number of point cloud pairs in candidate pose hypotheses that do not contain the correct transformation, $K$ denotes the number of candidate poses for each point cloud pair, $S$ represents the confidence scores predicted by the pose evaluator, and $\lambda$ is the truncation threshold. The metric indicates the ratio of point cloud pairs, among the total, for which the confidence predictions of all $K$ candidate poses are below the threshold $\lambda$.

\section{More Implementation Details}
\subsection{Datasets}
{\noindent \textbf{Indoor scenes.}}
We use the 3DMatch benchmark \cite{Zeng20163DMatchLL} for evaluating the performance on indoor scenes. It contains 1623 pairs of point clouds with ground-truth camera poses, which are obtained by 8 different RGBD sequences. For each pair of point clouds, we downsample them using a voxel size of 5cm. Then, we use different descriptors and pose estimators to extract correspondences and candidate poses, respectively. Partial overlapping is challenging in point cloud registration. In order to further test the performance of our method, 3DLoMatch benchmark \cite{huang2021predator} is adopted to further verify the performance of the algorithm on low-overlapped point cloud registration. It contains 1781 pairs of point clouds with overlap ratios ranging from 10\% to 30\%.

{\noindent \textbf{Outdoor scenes.}} We use the KITTI benchmark~\cite{Geiger2012AreWR} for evaluating the performance on outdoor scenes. It contains 11 outdoor driving scenarios of point clouds. Following \cite{Choy2019FullyCG}, we choose the 8 to 10 scenarios as test datasets. For all the LIDAR scans, we use the first scan that is taken at least 10cm apart within each sequence to create a pair, which can obtain 555 pairs of point clouds for testing.

{\noindent \textbf{Multiway registration scenes.}} We use the ICL-NUIM augments  benchmark \cite{Handa2014ABF} for evaluating the performance on multiway scenes. It consists of four camera trajectories from two scenes for testing. Following \cite{Bai2021PointDSCRP, chen2022sc2}, we fuse 50 consecutive RGBD frames to generate the point cloud fragements. 

\subsection{Network Architecture Details}
\noindent{\textbf{Feature Extractor.}}
We utilized the pre-trained GeoTransformer as the backbone network for feature extraction. Notably, to achieve a lightweight design, we only utilize two layers of points and their corresponding features. Specifically, for the 3DMatch and KITTI datasets, we selected the coarsest layer and the penultimate layer, with feature dimensions of 256 and 512, respectively. For detailed network architecture, please refer to the original implementation. To further learn the features and facilitate subsequent computations, we employed two linear layers, Linear1 (256->256) and Linear2 (512->256), to map the obtained features to the same dimension. Specifically, the points and their updated features at the coarsest layers are represented as $\hat{\mathcal{P}}$/$\hat{\mathcal{Q}}$ and $\hat{\mathbf{F}}^{\mathcal{P}}$/$\hat{\mathbf{F}}^{\mathcal{Q}} \in \mathbb{R}^{|\hat{\mathcal{P}}/\hat{\mathcal{Q}}| \times d}$, while  those at the penultimate layers are represented as $\Tilde{\mathcal{P}}$/$\Tilde{\mathcal{Q}}$ and $\Tilde{\mathbf{F}}^{\mathcal{P}}$/$\Tilde{\mathbf{F}}^{\mathcal{Q}} \in \mathbb{R}^{|\Tilde{\mathcal{P}}/\Tilde{\mathcal{Q}}| \times d}$ ($d=256$).

\begin{figure*}[h]
  \centering
  \includegraphics[width=0.7\linewidth]{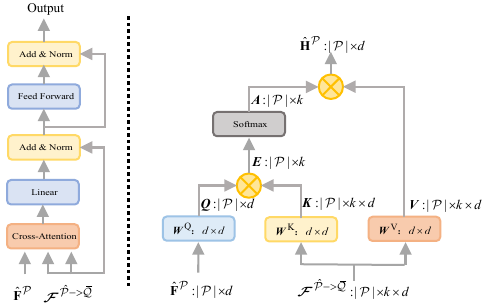}
  \caption{Left: The structure of pose-aware attention module.
    Right: The computation graph of pose-aware attention..
  }
  \label{fig.att}
\end{figure*}

\noindent{\textbf{Pose-Aware Attention Module.}}
We adopted a multi-head attention mechanism with the number of attention heads set to 4. We flexibly adjusted the attention regions based on different candidate poses to obtain the corresponding feature volume ${\mathcal{F}}^{\hat{\mathcal{P}} \rightarrow  \Tilde{\mathcal{Q}}} \in \mathbb{R}^{|\hat{\mathcal{P}}| \times k \times d}$ for $\hat{\mathbf{F}}^{\mathcal{P}}$. The computation of the pose-aware attention features $\hat{\mathbf{H}}^{{\mathcal{P}}}$ for $\hat{\mathbf{F}}^{\mathcal{P}}$ is shown in \cref{fig.att}, and the $ \hat{\mathbf{H}}^{{\mathcal{Q}}}$ is computed in the same way.

\begin{figure}[h]
  \centering
  \includegraphics[width=.8\linewidth]{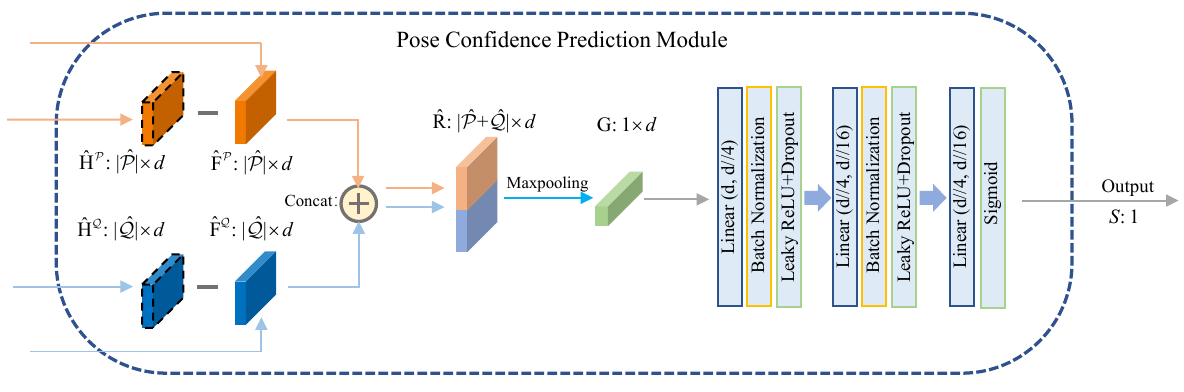}
  \vspace{-10pt}
  \caption{The structure of Pose Confidence Prediction Module.
  }
  \label{fig.pcp}
\end{figure}

\noindent{\textbf{Pose Confidence Prediction Module.}}
As shown in \cref{fig.pcp}, to capture the differentiation in feature updates under different poses, we apply a residual operation to the obtained features in this context: $\hat{\mathbf{R}}^{{\mathcal{P}}/{\mathcal{Q}}} = \hat{\mathbf{H}}^{{\mathcal{P}}/{\mathcal{Q}}} - \hat{\mathbf{F}}^{{\mathcal{P}}/{\mathcal{Q}}}$. Then we concatenate the obtained residual features: $\textbf{Cat}(\hat{\mathbf{R}}^{\mathcal{P}},\hat{\mathbf{R}}^{\mathcal{Q}}) \in \mathbb{R}^{|\mathcal{P} + \mathcal{Q}| \times d}$  and then perform a max-pooling operation to obtain the global feature: $\mathbf{G} \in \mathbb{R}^{1 \times d}$. Subsequently, we employ a 3-layer MLP with 256->64->16->1 channels to map and obtain the corresponding confidence scores $S$. Following the first two linear layers, each layer is succeeded by 1D batch normalization, a leaky ReLU, and a dropout layer. Since we consider the problem as binary classification, we apply the sigmoid layer after the final linear layer.

\subsection{Training Details}
We train indoor and outdoor Deep-PE models on 3DMatch and KITTI datasets. To generate training and validation sets, we first use pre-trained indoor and outdoor Geotransformer~\cite{Qin2022GeometricTF} models to extract correspondences. We then use SC2-PCR~\cite{chen2022sc2} as the pose estimator to extract 1000 candidate poses for each point cloud pair. In this scenario, we observed instances where correct or incorrect poses were absent. To ensure an adequate presence of both correct and incorrect poses in the candidate pose pool, we introduced perturbations by randomly sampling small rotations ($Rot\in [0^{\circ}, 15^{\circ}]$ for 3DMatch, $Rot\in [0^{\circ}, 5^{\circ}]$ for KITTI) or large rotations ($Rot\in [15^{\circ}, 60^{\circ}]$ for 3DMatch, $Rot\in [5^{\circ}, 20^{\circ}]$ for KITTI) and multiplying them with the ground truth relative pose. This process aims to augment the pool with a diverse set of correct and incorrect poses. For each candidate pose, we calculate the corresponding Root Mean Square Error (RMSE) according to the ground-truth correspondences.

\subsection{Parameters Details}
We implement and evaluate our Deep-PE with PyTorch on an AMD EPYC 7642 and an NVIDIA RTX 3090 GPU. The network is trained with the Adam optimizer for 40 epochs on 3DMatch and 80 epochs on KITTI. The batch size is 20, and the weight decay is $1 \times 10^{-5}$. The learning rate starts from $1 \times 10^{-6}$ and decays exponentially by 0.05 every epoch on 3DMatch and every 4 epochs on KITTI. In the feature extractor module, the linear layer maps uniformly to dimension $d=256$. In the pose-aware attention module, the number of nearest neighbors $k$ is set to 16, and the threshold $t$ is set to 0.1. For the parameters of the loss function, $\alpha$, $\beta$, and $\gamma$ are set to 5, 0.2, and 2, respectively, ensuring that the weight terms are within the range of 0 to 1. In the pose confidence prediction module, the parameter of the dropout layer is set to 0.5. In the actual test, to improve the speed of pose evaluation of our model, CC~\cite{Fischler1981RandomSC} is used to preprocess the candidate pose $\mathcal{H}$, and $|\mathcal{H}|* \delta$ candidate poses are selected according to the statistical scores. Based on our experience, we have found that setting $\delta=0.4$ achieves a balance between performance and speed.

\section{Additional Experiments}
\subsection{Detailed Results on Indoor Scenes}
\noindent{\textbf{Efficiency with different $\delta$.}} 
The advantage of statistics-based pose evaluators over our method lies in computational efficiency, prompting exploration into combining the two approaches. In our model, the ideal scenario involves minimizing the number of candidate poses while ensuring the presence of at least one correct transformation. Therefore, we employ the statistics-based pose evaluator CC \cite{Fischler1981RandomSC} for candidates' pre-processing, preventing unnecessary calculations.
To provide a deeper understanding of Deep-PE, we present the running time and registration recall results for different values of $\delta$ in \cref{tab.B}. As the value of $\delta$ increases, more candidate poses can be retained, but this comes at the expense of increased computational time. Notably, even when $\delta>0.2$, although this introduces a significant number of incorrect candidate poses, our method maintains stable performance. Setting $\delta$ to 0.4 (about 400 candidate poses) achieves a balanced trade-off between efficiency and accuracy.

\begin{table}[h]
\centering
\caption{The running time and registration recall with different $\delta$.}
\setlength{\tabcolsep}{12pt}
\scriptsize
\centering
\label{tab.B}
\begin{tabular}{cccccccc} 
\toprule
             & \multicolumn{3}{c}{3DMatch RR(\%)$\uparrow$}     & \multicolumn{3}{c}{3DLoMatch RR(\%)$\uparrow$}   & \multicolumn{1}{c}{Time(s)$\downarrow$}  \\ 
\cmidrule(l){2-4} \cmidrule(l){5-7}
             & FPFH           & FCGF           & GEO            & FPFH           & FCGF           & GEO            &                                           \\ 
\midrule
$\delta$=0.1 & 86.26          & 94.39          & 95.81          & 47.00          & 68.39          & 78.81          & 0.19                                      \\
$\delta$=0.2 & 88.42          & 94.70          & 95.81          & 50.00          & 70.75          & 79.28          & 0.34                                      \\
$\delta$=0.3 & \uline{88.54}  & \textbf{95.07} & 95.63          & 50.53          & 71.25          & 79.28          & 0.44                                      \\
$\delta$=0.4 & \textbf{89.40} & \uline{94.82}  & 95.93          & 50.25          & 71.81          & \uline{80.23}  & 0.61                                      \\
$\delta$=0.5 & 88.97          & 94.58          & \uline{96.12}  & 50.42          & \uline{72.26}  & 80.01          & 0.67                                      \\
$\delta$=0.6 & 89.40          & 94.52          & \uline{96.12}  & 50.53          & \textbf{72.71} & 79.79          & 0.82                                      \\
$\delta$=0.7 & 89.09          & 94.33          & 96.06          & \uline{50.65}  & 71.81          & 79.79          & 0.92                                      \\
$\delta$=0.8 & 89.34          & 94.52          & 95.93          & 50.31          & 71.87          & 79.90          & 1.07                                      \\
$\delta$=0.9 & 88.97          & 94.58          & \uline{96.12}  & 50.31          & 72.09          & 79.73          & 1.16                                      \\
$\delta$=1.0 & 89.28          & 94.58          & \textbf{96.24} & \textbf{50.87} & 72.21          & \textbf{80.24} & 1.32                                      \\
\bottomrule
\end{tabular}
\vspace{-8pt}
\end{table}

\begin{figure*}[h]
  \centering
  \includegraphics[width=\linewidth]{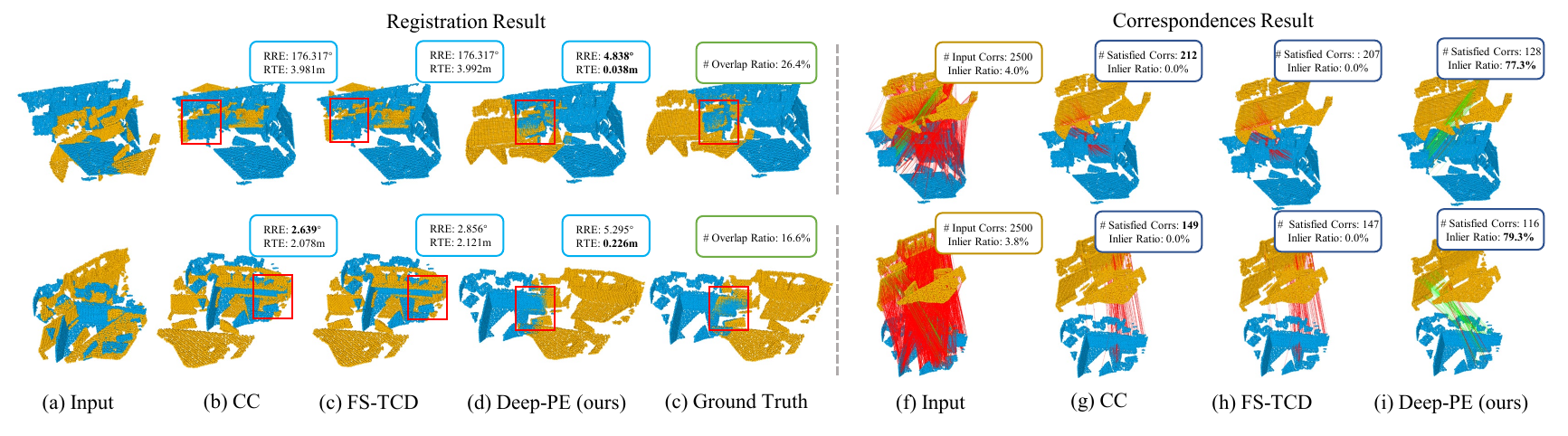}
  \vspace{-20pt}
  \caption{Comparison of registration results between Deep-PE and statistics-based pose evaluators (Under FPFH).
  }
  \label{fig.10}
\end{figure*}

\begin{figure*}[h]
  \centering
  \includegraphics[width=\linewidth]{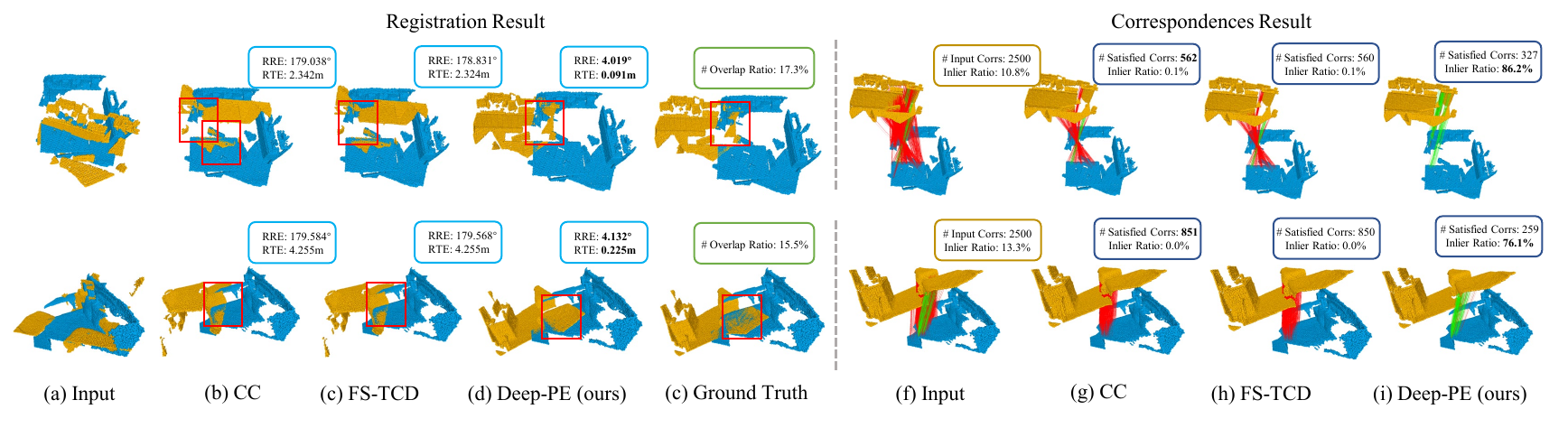}
  \vspace{-20pt}
  \caption{Comparison of registration results between Deep-PE and statistics-based pose evaluators (Under Geotransformer).
  }
  \label{fig.11}
\end{figure*}

\noindent{\textbf{More Qualitative Results.}} 
We provide more qualitative results on 3DLoMatch in \cref{fig.10} and \cref{fig.11}. We use handcrafted FPFH \cite{Rusu2009FastPF} and the advanced Geotransformer (GEO) \cite{Qin2022GeometricTF} extract correspondences respectively, we compared our method to the commonly used CC \cite{Fischler1981RandomSC} and the advanced FS-TCD~\cite{chen2023sc}. We reported the visual registration and correspondence results in challenging scenarios. When the inlier ratio of input correspondences is low, statistics-based  evaluation mechanisms become ineffective, leading to registration failures. In contrast, our method incorporates global alignment priors and utilizes a carefully designed neural network to learn and evaluate the quality of pose alignment. Consequently, it consistently identifies the optimal pose among the candidates, even if it does not satisfy the majority of correspondences.

\begin{table}[h]
\centering
\caption{Registration results with different pose evaluators on KITTI.}
\setlength{\tabcolsep}{10pt}
\scriptsize
\centering
\label{tab.C}
\begin{tabular}{lcccccc} 
\toprule
                       & \multicolumn{3}{c}{FPFH}                                    & \multicolumn{3}{c}{FCGF}                                     \\ 
\cmidrule(l){2-4} \cmidrule(l){5-7}
                       & RR(\%)$\uparrow$ & RRE(deg)$\downarrow$ & RTE(cm)$\downarrow$ & RR(\%)$\uparrow$ & RRE(deg)$\downarrow$ & RTE(cm)$\downarrow$  \\ 
\midrule
CC \cite{Fischler1981RandomSC}      & \textbf{99.64}                              & 0.36          & 7.89          & 98.20          & \textbf{0.33} & 20.89           \\
MAE \cite{Yang2021TowardEA}     & \textbf{99.64}                              & 0.36          & 7.81          & 98.20          & \textbf{0.33} & 20.78           \\
MSE \cite{Yang2021TowardEA}     & 99.34                              & \textbf{0.35} & \uline{7.73}  & 98.20          & \textbf{0.33} & \uline{20.76}   \\
FS-TCD \cite{chen2023sc}   & \textbf{99.64} & \textbf{0.35} & 7.74          & \uline{98.56}  & \textbf{0.33} & 20.85           \\
Deep-PE (\textit{ours}) & \textbf{99.64}                     & \textbf{0.35} & \textbf{7.68} & \textbf{98.72} & \textbf{0.33} & \textbf{20.74}  \\
\bottomrule
\end{tabular}
\vspace{-7pt}
\end{table}

\subsection{Evaluation on Outdoor Scenes}
In \cref{tab.C}, we compare our method with four statistics-based pose evaluators, including Correspondences Counting (CC)~\cite{Fischler1981RandomSC}, Mean Average Error (MAE)~\cite{zhang2023MAC, Yang2021TowardEA}, Mean Square Error (MSE)~\cite{Yang2021TowardEA}, and Feature and Spatial consistency constrained Truncated Chamfer Distance (FS-TCD)~\cite{chen2023sc}. It's important to note that SC2-PCR~\cite{chen2022sc2} is utilized here to generate candidate poses, and different pose evaluators are employed to select the final transformation. As shown in the table, Deep-PE consistently delivers the best results across all metrics and shares the top spot with FPFH~\cite{Rusu2009FastPF} and FCGF~\cite{Choy2019FullyCG} descriptor settings, respectively.

\subsection{Evaluation on Multiway Registration Scenes}

To evaluate multi-view registration, we used the enhanced ICL-NUIM dataset \cite{Handa2014ABF}. In the experimental setup, we employed handcrafted descriptors FPFH \cite{Rusu2009FastPF} and  advanced Geotransformer \cite{Qin2022GeometricTF} to extract correspondences, followed by using SC2-PCR \cite{chen2022sc2} as the pose estimator to generate a set of candidate poses. Here, we compared our method with all known pose evaluators. It is important to note that, to test the model's generalization ability, we directly used the model trained on 3DMatch without retraining or fine-tuning. As shown in \cref{tab.D}, due to the performance limitations of the FPFH \cite{Rusu2009FastPF} descriptor, statistics-based pose evaluators are always more significantly impacted. Although FS-TCD \cite{chen2023sc} establishes more correspondences and uses Chamfer distance to evaluate global alignment, it remains overly dependent on the quality of the input correspondences. Since our method fully simulates and learns the alignment state of point clouds and performs confidence prediction through neural network, it is sensitive only to the pose and not reliant on the quality of the input correspondences. Thus, our method achieves the best performance on all scenarios. Under Geotransformer \cite{Qin2022GeometricTF}, our method still maintains certain advantages across multiple scenarios. It is evident that our method demonstrates strong generalization ability in unknown and more complex scene applications.

\begin{table}[h]
\centering
\caption{Registration results on Augmented ICL-NUIM.}
\label{tab.D}
\setlength{\tabcolsep}{10pt}
\scriptsize
\centering
\begin{tabular}{lccccc} 
\toprule
        & \multicolumn{5}{c}{FPFH ATE (cm)$\downarrow$}                                                           \\ 
\cmidrule{2-6}
        & Living1        & Living2        & Office1        & Office2       & Mean            \\ 
\midrule
CC \cite{Fischler1981RandomSC}      & 18.68          & 14.31          & 14.63          & 11.95         & 14.90           \\
MAE \cite{Yang2021TowardEA}     & 18.36          & 14.24          & 14.48          & 11.82         & 14.73           \\
MSE \cite{Yang2021TowardEA}     & 18.24          & 14.12          & 14.32          & 11.76         & 14.61           \\
FS-TCD \cite{chen2023sc} & 17.56          & 14.37          & 13.24          & 9.49          & \uline{13.67}   \\
Deep-PE (\textit{ours}) & \textbf{17.32} & \textbf{14.02} & \textbf{12.94} & \textbf{9.28} & \textbf{13.39}  \\ 
\midrule
        & \multicolumn{5}{c}{Geotransformer ATE (cm)$\downarrow$}                                                  \\ 
\cmidrule{2-6}
        & Living1        & Living2        & Office1        & Office2       & Mean            \\ 
\midrule
CC \cite{Fischler1981RandomSC}     & 17.48          & 15.26          & 13.80          & 9.72          & 14.07           \\
MAE \cite{Yang2021TowardEA}     & 17.44          & 15.18          & 13.78          & 9.66          & 14.02           \\
MSE \cite{Yang2021TowardEA}     & 17.40          & \uline{15.02}  & 13.78          & 9.52          & 13.93           \\
FS-TCD \cite{chen2023sc}  & \textbf{17.22} & 15.16          & \uline{13.42}  & \textbf{9.31} & \uline{13.78}   \\
Deep-PE (\textit{ours}) & \uline{17.26}  & \textbf{14.62} & \textbf{13.29} & \uline{9.48}  & \textbf{13.67}  \\
\bottomrule
\end{tabular}
\end{table}

\begin{table}[h]
\centering
\caption{Loss supervision ablation results on the 3DMatch and 3DLoMatch dataset.}
\setlength{\tabcolsep}{10pt}
\scriptsize
\centering
\begin{tabular}{lcccccc} 
\toprule
                       & \multicolumn{3}{c}{3DMatch}                                                 & \multicolumn{3}{c}{3DLoMatch~ ~}                                             \\ 
\cmidrule(l){2-4} \cmidrule(l){5-7}
                       & RR(\%) $\uparrow$         & \multicolumn{1}{l}{RRE(deg) $\downarrow$} & \multicolumn{1}{l}{RTE(cm) $\downarrow$} & RR(\%) $\uparrow$    & \multicolumn{1}{l}{RRE(deg) $\downarrow$} & \multicolumn{1}{l}{RTE(cm) $\downarrow$}  \\ 
\midrule
L1(R,t)                & 82.78          & 2.49                         & 6.96                        & 37.29          & 4.66                         & 12.22                        \\
L1(RMSE)               & 85.52          & \uline{2.32}                 & \uline{6.77}                & 44.14          & \uline{4.57}                 & \uline{11.78}                \\
Cross-Entropy          & \uline{87.12}  & 2.78                         & 7.23                        & \uline{48.66}  & 4.96                         & 12.32                        \\
Weighted Cross-Entropy & \textbf{89.40} & \textbf{2.18}                & \textbf{6.52}               & \textbf{50.25} & \textbf{4.32}                & \textbf{11.24}               \\
\bottomrule
\end{tabular}
\label{tab.E}
\vspace{-7pt}
\end{table}

\subsection{Ablation Study of Loss Function}
As illustrated in \ref{tab.E}, we present the experimental results of training Deep-PE using various supervisory signals (e.g., FPFH~\cite{Rusu2009FastPF} as the descriptor, SC2-PCR~\cite{chen2022sc2} for generating candidate poses). In this context, the candidate supervisory signals encompass L1(R, t), L1(RMSE), cross-entropy, and the weighted cross-entropy proposed in this paper. Notably, due to substantial variations in pose differences within candidate poses, we observed that directly employing the L1 loss function to train the network to predict rotation and translation errors, or RMSE, presented significant training challenges. The latter, as it predicts a single value, performed slightly better than the former. Treating the problem as a binary classification task yielded superior results, and our proposed weighted cross-entropy, which places emphasis on the pose with the smallest error, achieved optimal performance on both the 3DMatch and 3DLoMatch datasets.

\subsection{Ablation Study of Modules}
In this section, we carry out comprehensive ablation studies to gain a deeper understanding of the different modules within our approach. We employ Geotransformer~\cite{Qin2022GeometricTF} to generate correspondences on both 3DMatch and 3DLoMatch datasets, while SC2-PCR~\cite{chen2022sc2} with CC~\cite{Fischler1981RandomSC} serves as the baseline, as depicted in Row (1) of \cref{tab.F}.

\begin{table}[h]
\caption{Ablation experiments of main modules on 3DMatch and 3DLoMatch.}
\setlength{\tabcolsep}{7pt}
\scriptsize
\centering
\label{tab.F}
\begin{tabular}{lcccccccccc} 
\toprule
                     & \multicolumn{1}{l}{} & \multicolumn{1}{l}{} & \multicolumn{1}{l}{} & \multicolumn{1}{l}{} & \multicolumn{3}{c}{3DMatch}                    & \multicolumn{3}{c}{3DLoMatch}                    \\ 
\cmidrule(l){6-8} \cmidrule(l){9-11}
\multicolumn{1}{c}{} & PT                   & GEO                  & PAA                  & PCP                  & RR(\%) $\uparrow$         & RRE(deg) $\downarrow$     & RTE(cm) $\downarrow$       & RR(\%) $\uparrow$         & RRE(deg) $\downarrow$     & RTE(cm) $\downarrow$         \\ 
\midrule
(1)                  & $\times$             & $\times$             & $\times$             & $\times$             & 83.98          & 2.24          & 6.80          & 38.57          & \textbf{4.16} & \textbf{10.23}  \\
(2)                  & $\times$             & $\checkmark$         & $\checkmark$         & $\checkmark$         & -              & -               & -              & -              & -              & -                \\
(3)                  & $\checkmark$         & $\times$             & $\checkmark$         & $\checkmark$         & \uline{87.92}  & \textbf{2.12} & \uline{6.69}  & \uline{47.45}  & 4.27          & 10.92           \\
(4)                  & $\checkmark$         & $\checkmark$         & $\times$             & $\checkmark$         & 81.21          & 2.34          & 6.92          & 33.18          & 4.89          & 10.78           \\
(5)                  & $\checkmark$         & $\checkmark$         & $\checkmark$         & $\times$             & 86.63          & 2.22          & 6.75          & 44.47          & \uline{4.22}  & \uline{10.37}   \\
(6)                  & $\checkmark$         & $\checkmark$         & $\checkmark$         & $\checkmark$         & \textbf{89.40} & \uline{2.18}  & \textbf{6.52} & \textbf{50.25} & 4.32          & 11.24           \\
\bottomrule
\end{tabular}
\vspace{-7pt}
\end{table}

\noindent{\textbf{Feature Extractor with Pre-Train. }}
We utilize the pre-trained Geotransformer~\cite{Qin2022GeometricTF} model as our feature extraction module and refrain from participating in subsequent network updates. This choice is motivated by our aim to maintain a stable point-wise feature representation of the input point clouds. Nevertheless, we also attempted to train Deep-PE in an end-to-end manner; however, the network did not converge, as shown in Row (2). Our analysis indicates that the loss function used solely for supervising pose confidence does not effectively guide the generation of point-wise features. Furthermore, due to limitations in our network structure, certain loss functions for supervising feature matching, such as circle loss~\cite{Qin2022GeometricTF}, could not be effectively adapted. As mentioned earlier, for the feature extraction module, we opt for the pre-training mode over the end-to-end approach.

\noindent{\textbf{Performance of Feature Extractor. }} 
To investigate the influence of feature description quality on our results, we conducted an experiment where we replaced the geometric self-attention module in Geotransformer~\cite{Qin2022GeometricTF} with the vanilla self-attention module~\cite{Vaswani2017AttentionIA}, keeping other modules unchanged. We then retrained the model with this modification. As depicted in Row (3) and Row (6), the RR obtained using the geometric self-attention module is 1.48\% higher than that achieved with the vanilla self-attention module on the 3DMatch dataset and 2.80\% higher on the 3DLoMatch dataset. This indicates that improved feature description can result in enhanced performance of our method.

\noindent{\textbf{Pose-Aware Attention Module. }} 
Being the most pivotal module in our approach, feature updating is accomplished through local attention operations under specific poses, resulting in distinct feature representations for different poses. However, during module ablation experiments, all feature representations become independent of pose, introducing ambiguity and hindering network convergence. To underscore the significance of this module, we conducted an experiment where we directly sought nearest neighbors under different poses to generate feature residuals, serving as a replacement for the attention module.
The results, as shown in the complete model (Row 6) compared to Row (4), reveal that the RR obtained using the pose-aware attention module is 8.19\% higher than when using the direct feature residual approach on the 3DMatch dataset and 17.07\% higher on the 3DLoMatch dataset. This underscores that the module effectively learns the degree of feature matching under different poses through an attention mechanism, rather than relying solely on simple residuals.

\noindent{\textbf{Pose Confidence Prediction Module. }} 
We utilize a single linear layer and a sigmoid layer to directly process the learned feature residuals, deviating from the design in this paper. Comparing Row (5) and Row (6), it becomes evident that employing a multi-layer perceptron for additional feature learning enhances the RR by 2.77\% on the 3DMatch dataset and 5.78\% on the 3DLoMatch dataset. This underscores the importance of further refining the acquired feature residuals through additional learning.

\section{Limitations}
Deep-PE relies on the candidate poses generated by the pose estimator and then finds the correct transformation from them, replacing the existing statistics-based pose evaluation mechanism. However, when no correct transformations exist in the candidate poses, our method cannot generate a correct transformation, which is the upper bound of our method. 

\begin{figure}[h]
  \centering
  \includegraphics[width=0.7\linewidth]{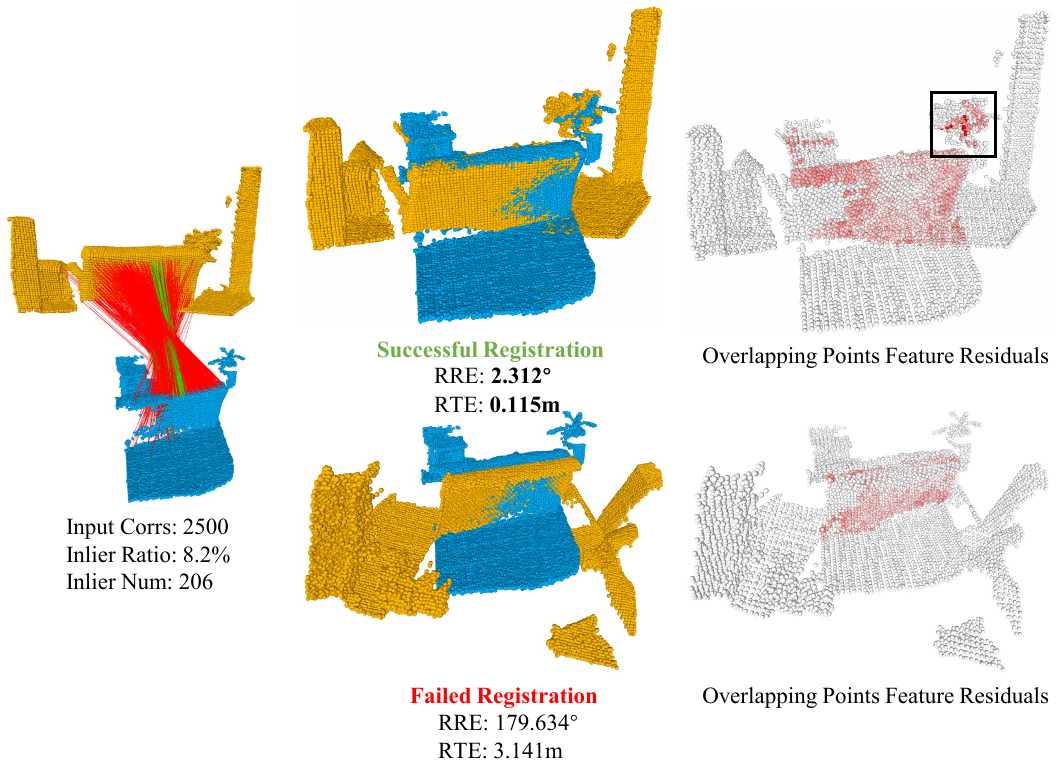}
  \caption{A failure example. For the input correspondences, it can be observed that a substantial number of correspondences are established in planar regions, indicating a high degree of similarity in features within this area. Under the correct pose, the upper-right corner, highlighted by the black box, exhibits poor feature description quality due to partial-missing, resulting in the inability to establish effective correspondences. Consequently, significant feature disparities arise during the alignment process. This region is located in the overlapping section, and compared to the correct pose, the incorrect pose aligns only the planar region. In this scenario, the incorrect pose paradoxically attains a higher confidence score. }
  \label{fig.12}
\end{figure}

In addition, we present a failure case, as illustrated in Fig.~\ref{fig.12}. Under the correct pose, the point cloud in the upper right corner exhibits severe omissions, resulting in blurred features and the failure to establish effective correspondences. However, this region represents the overlapping area under the correct pose, where the features of the planar region points exhibit minimal differences, leading to the establishment of numerous correspondences. Due to our model considering only the pose with overlapping planar regions as the optimal transformation, this scenario is unavoidable.

\end{document}